%% file: neurips_2025.tex
\documentclass{article}

\usepackage[preprint]{neurips_2026}
\usepackage[utf8]{inputenc} 
\usepackage[T1]{fontenc}    
\usepackage{url}            
\usepackage{booktabs}       
\usepackage{amsfonts}       
\usepackage{nicefrac}       
\usepackage{microtype}      

\usepackage{xcolor}          
\definecolor{accent}{HTML}{1F4E79}   
\definecolor{extern}{HTML}{7D4E2D}   

\usepackage{hyperref}
\hypersetup{
  colorlinks   = true,
  allcolors    = accent,
  urlcolor     = extern,
  linktoc      = all,
  breaklinks   = true,
  bookmarksnumbered = true,
}
\usepackage{todonotes}
\usepackage{graphicx}
\usepackage{subcaption}
\usepackage{tikz}
\usetikzlibrary{positioning,fit,arrows.meta,calc}
\usepackage{float}
\usepackage{amsmath}
\usepackage{amssymb}
\usepackage{mathtools}
\usepackage{amsthm}
\usepackage{placeins}
\usepackage[nameinlink,capitalise]{cleveref}  

\title{Amortized Structured Stochastic Variational Inference
for Gaussian Process Latent Variable Models}

\author{
\textbf{
Maksym Tretiakov$^{1,2}$,
Sarah Lucie Filippi$^{3}$,
Vincent Fortuin$^{2,4,5}$,
Ruth Misener$^{3}$
}
\\[-0.1em]
\textbf{
Ruby Sedgwick$^{6}$,
James Odgers$^{2,4,5}$
}
\\[0.5em]
{\fontseries{m}\selectfont
$^{1}$LMU Munich,
$^{2}$Munich Center for Machine Learning (MCML),
$^{3}$Imperial College London}
\\[-0.1em]
{\fontseries{m}\selectfont
$^{4}$University of Technology Nuremberg (UTN),
$^{5}$Helmholtz AI,
$^{6}$Xyme}
}

\theoremstyle{plain}

\theoremstyle{definition}

\theoremstyle{remark}

\newcommand{\R}{\mathbb{R}}
\newcommand{\I}{\mathbb{I}}
\newcommand{\elbo}{\mathcal{L}}

\newcommand{\KL}{D_{KL}}
\newcommand{\fit}{\mathcal{F}}

\begin{document}

\maketitle

\begin{abstract}
    Many machine learning methods aim to approximate the lower-dimensional manifold on which the data lives. A desirable feature of such methods is that they should capture the epistemic uncertainty of this learned manifold. One model that achieves this is the Gaussian Process Latent Variable Model, in which a Gaussian Process (GP) mapping from the latent space provides an estimate of the uncertainty of the manifold. However, the effectiveness of this uncertainty estimation is limited by the mean-field variational approximation between the GP inducing points and the latent variables. In this work, we apply Amortized Structured Stochastic Variational Inference to allow the variational posterior for the latent space to be conditionally dependent on the value of the inducing points. We demonstrate that this more flexible variational posterior improves several metrics relating to the reconstruction of points on the data manifold. 
\end{abstract}


\section{Introduction}
\label{sec:intro}

In many real-world settings, observed high-dimensional data has a complex structure that can be naturally explained by a smaller number of unobserved factors. 
This general formulation is relevant for a broad class of models, starting from classical models, such as Probabilistic Principal Component Analysis \citep[PPCA;][]{tipping1999ProbabilisticPrincipalComponent}, to modern deep generative models, such as Variational Auto-Encoders \citep[VAEs;][]{Kingma2014}. These approaches can uncover interpretable low-dimensional structure and enable downstream inference tasks.  

Unfortunately, in many practical tasks where learning a data manifold is desirable, the data is likely to be both noisy and limited in number. In such cases, it is important to learn not only point estimates for the data manifold, but also to estimate the epistemic uncertainty of the manifold. One model that can achieve this is the Gaussian Process Latent Variable Model \citep[GPLVM;][]{lawrence2005probabilistic, titsias2010bayesian, damianou2016variational}. Thanks to its non-parametric Bayesian approach, this method can both learn arbitrarily complex data manifolds and remain robust to overfitting. These properties have allowed the GPLVM and its variants to be applied to a diverse set of applications, including biology \citep{Reid2016, Ahmed2019, sedgwick2025TransferlearningBayesian}, chemical engineering \citep{odgers2025WeightedSumGaussian}, causal learning \citep{dhir2024BivariateCausalDiscovery, dhir2025ContinuousBayesianModel}, and robotics \citep{jaquier2024Bringingmotiontaxonomies, rozo2025Riemann$^2$LearningRiemannian}.

However, all approaches to the GPLVM share a shortcoming: a Mean Field (MF) approximation in the variational family. Specifically, current methods approximate the posterior of the positions of the data in the latent space and the mapping from the latent space to the observations to be independent of one another \citep{lalchandGeneralisedGaussianProcess2022}. This approach results in the model underestimating the epistemic uncertainty in the GP mapping. In this work, we apply Structured Stochastic Variational Inference \citep[SSVI;][]{hoffman2015ssvi} to introduce dependence of the latent variable on the Sparse GP inducing points, thereby improving the approximation to the true posterior. 

The rest of the paper is laid out as follows: first, in Section~\ref{sec:background}, we provide an overview of the background required for this paper. Following this, in Section~\ref{sec:method} we introduce our method and derive the Evidence Lower Bound (ELBO) and discuss the relationship of our approach to existing methods in Section~\ref{sec:related-works}.  In Section~\ref{sec:experiments}, we demonstrate that our method outperforms the existing MF methods and argue that it provides a better estimate of the epistemic uncertainty. Finally, in Sections~\ref{sec:limitations} and \ref{sec:conclusion}, we discuss some of the current limitations of our approach and suggest further work which could take advantage of our variational family.   

\section{Background}
\label{sec:background}
The problem of learning the latent structure from high-dimensional data has been studied for a significant amount of time \citep{tipping1999ProbabilisticPrincipalComponent}. 
We follow a standard setup by assuming that we have high-dimensional observations, $y_n \in \R^P$, which we seek to describe by a lower-dimensional latent variable, $x_n \in \R^Q$, where $P \gg Q $. We assume that there exists a mapping $f: \R^Q \rightarrow \R^P$, such that the relationship between $x_n$ and $y_n$ is 
\begin{equation}
    y_n = f(x_n) + \epsilon_n, 
\end{equation}
with $\epsilon_n \in \R^P$ representing random observation noise. For the duration of this paper, we assume it follows the distribution $\epsilon_n \sim N(0, \sigma^2 \I_P) $, and place a prior on $x_n \sim N(0,\I_Q)$. For notational convenience, we will also use $f_n = f(x_n) \in \R^{P}$.  

We assume that we have a large number of high-dimensional data points,  $Y \in \R^{N \times P}$ which are associated with values for the latent variables, $X \in \R^{N \times Q}$, and function values, $F \in \R^{N \times P}$. The matrices are constructed such that $x_n$, $f_n$, and $y_n$ are the $n$th row of $X$, $F$, and $Y$ respectively. We assume that we have observed $Y$, and that the goal of our probabilistic model is to infer probability distributions for the possible values of $X$ and $F$.

We take a Bayesian approach in this paper, ideally aiming to learn a posterior distribution $p(F,X|Y)$. This requires us to specify a prior for both the function and the latent variables.  While the prior over the latent space can be used to encode useful information \citep{damianou2016variational}, we will use an independent spherical  Normal prior for each $x_n$, so
\begin{equation}
    p(x_n) =  N(0, \I_Q).
\end{equation}

Instead of attempting to learn a parametric representation of $f(\cdot)$, we model it using Gaussian Processes (GPs). GPs are a non-parametric Bayesian technique in which each finite set of function values is assumed to follow a Multivariate Normal Distribution \citep{rasmussen_gaussian_2008}. We apply an independent zero-mean GP prior over each of the output dimensions of our high-dimensional observation, \textit{i.e.}, we take  
$f_{p} \in \R^{N}$ to be the $p$th column of $F$ and use the prior  
\begin{equation}
    p(f_p | X) = N(0, K_{XX}).
\end{equation}
The covariance matrix, $K_{XX}$ is given by a kernel function, $[K_{XX}]_{ij} = k(x_i, x_j)$, which controls properties of the function and can be fit to data using the marginal likelihood \citep{rasmussen_gaussian_2008}, which, following \citet{titsias2010bayesian}, we assume to be an Automatic Relevance Determination (ARD) kernel for this paper.

With these priors, we can  factorize the model as
\begin{equation}
    p(Y,F,X) = \prod_{n=1}^N  p(y_n|f_{n}) p(x_n) \prod_{p=1}^P p(f_p| X ). 
\end{equation}
Ideally, a GPLVM is the posterior distribution for both the latent variables and the functions, \textit{i.e.} $p(X,F|Y)$. Unfortunately, this is not possible as it would require $p(Y) = \int p(Y|F)p(F|X)p(X) dX dF $, which requires an intractable integral over the inverse of the kernel matrix $K_{XX}^{-1}$. 
Furthermore, as evaluating the likelihood of any new combination of latent variables would require computing the same inverse matrix $K_{XX}^{-1}$, sampling-based MCMC approaches would be prohibitively computationally expensive.  When GPLVMs were first introduced by \citet{lawrence2005probabilistic}, the uncertainty of the position of the inputs $x_i$ was ignored, resulting in a MAP estimate for the latent space.

In order to approximate the uncertainty in the latent space, we need to use a Variational Inference (VI) approach. VI is a method in machine learning and statistics that approximates the posterior by minimizing the KL divergence between an approximate posterior, $q(\cdot)$,  and the true posterior \citep{bleiVariationalInferenceReview2016}. This is achieved by maximizing an objective known as the Evidence Lower Bound (ELBO).
To apply Variational Inference to GPs, it is necessary to augment the GP with a set of fake data points, known as \textit{inducing points} \citep{titsiasVariationalLearningInducing2009, hensmanGaussianProcessesBig2013}. These inducing points are drawn from the stochastic process but are not associated with any observation.
Mathematically, we augment each of the $P$ independent GP mappings from the latent space to one dimension of the output space, $f_p: \R^Q \rightarrow \R$, with $M$ additional points. Each output dimension's $M$ inducing points have inputs in the latent space $Z_p \in \R^{M \times Q} $ and random output variables $u_p \in \R^{M}$.  For practical reasons of saving compute, each of the $Z_p$ inputs is assumed to have a single value, so $Z = Z_1 = \dots = Z_P$.  For notational convenience, we also introduce $U \in \R^{P \times M }$, which is the matrix formed by stacking the set of the output vectors, $u_p$. 

As $u_p$ is drawn from the same GP prior as $f_p$, the augmented prior for each output dimension is 
\begin{equation*}
(u_{p1}, \dots, u_{pM}, f_{p1}, \dots , f_{pN})^\top\
\sim \mathcal N\!\left(0, K_{(XZ)(XZ)}\right).
\end{equation*}
\noindent where
\begin{equation}
\label{eq:block-kernel}
K_{(XZ)(XZ)} =
\begin{pmatrix}
K_{ZZ} & K_{ZX} \\
K_{XZ} & K_{XX}
\end{pmatrix}.
\end{equation}
Furthermore, it is possible to write down each of the $P$ conditional distributions for function values for each observed dimension, $f_p \in \R^N$, given a particular value of the inducing points.  This distribution is identical to the distribution of predictions in the standard GP, so
    \begin{equation} \label{eq:conditional_f}
        p(f_p|u_p, X) = N(\mu^{f}_p, \Sigma^{f}_p)
    \end{equation}
    where the values for the mean and covariance matrix are given by
    \begin{equation*}
    \mu^{f}_p = K_{XZ}K_{ZZ}^{-1}u_p
    \end{equation*}
    \begin{equation*}
    \Sigma^{f}_p = K_{XX} - K_{XZ}K_{ZZ}^{-1}K_{ZX}. 
    \end{equation*}
    This dependence has been exploited in various ways, notably by \citet{hensmanGaussianProcessesBig2013} to apply Stochastic Variational Inference (SVI) to GPs. 
    
    In the context of GPLVMs, it allows us to write a tractable ELBO which includes an approximation of the uncertainty in the latent space \citep{titsias2010bayesian}. In order to derive an ELBO, it is necessary to specify a variational family to approximate the posterior. To the best of our knowledge,  
    all prior work has made a simplifying assumption at this point: they have made a Mean Field (MF) restriction between the latent variable and the GP posterior, so the variational posterior factorizes as
    \begin{equation} \label{eq:MF-factorisation}
        q(F,X,U) = \prod_{p=1}^P p(f_p \mid X, u_p) q(u_p) \prod_{n=1}^N q(x_n).
    \end{equation}


    By noting that the contribution to the ELBO from the KL divergence between the variational posterior and prior of $p(f_p| u_p, X)$ is zero, as the two terms are equal \citep{titsias2010bayesian, damianou2016variational}, the ELBO for MF-GPLVM can be written as 
\begin{equation}
\begin{split}
    \label{eq:mfvi_elbo}
    \mathcal{L}_{MF} = &
\sum_{n=1}^{N}\bigg[
\mathbb{E}_{p(f_n| x_n, U)q(x_n) q(U)}\!\left[
        \log \left(p(y_n\mid f_n, x_n, U)\right)
\right] \\ &- \KL(q(x_n)|| p(x_n)) \bigg] 
- \KL(q(U)||p(U)).
\end{split}
\end{equation}
Typically, a normal distribution is used for all of the variational posterior distributions, so $q(u_p) = N(\mu_p, \Sigma_p)$ and $q(x_n) = N(\mu_n, \Sigma_n)$. The ELBO can either be optimized directly by maximizing this objective over the constants $\{\mu_p, \Sigma_p\}_{p=1}^P $ and $\{\mu_n, \Sigma_n\}_{n=1}^N$, or by using an amortized strategy in which the parameters of the latent variable variational posterior are defined as the output of a function of the observations, \textit{i.e.} $(\mu_n, \Sigma_n) = g(y_n)$ with $g: \R^P \to \R^Q\times \R^{Q\times Q}$, and the function $g(\cdot)$ being parametrized by a neural network which is learned during training \citep{lalchandGeneralisedGaussianProcess2022}.

The key issue with this variational posterior is that every likely value for every latent variable, $q(x_n)$,  is combined with every likely value of the inducing points, $q(U)$,  when evaluating the expected log likelihood of $y_n$. This enforces a trade-off between accurately representing the uncertainty of the GP mapping and data points in the latent space. As VI is generally mode-seeking \citep{minka2005Divergencemeasuresmessage}, it tends to underestimate the epistemic uncertainty of the data manifold.

\section{Structured Stochastic Variational Inference for GPLVMs}
\label{sec:method}

Although the MF-GPLVM provides a tractable ELBO, which has been shown to achieve a good approximation of the latent space, the MF assumption is a strong and incorrect one. In essence we are asserting that the latent variable posterior, $q(X)$, should not depend on the  mapping from the latent space to the observed space. This seems intuitively incorrect, and the more precise mathematical issue can be seen in the expression for $p(f_p| u_p, X)$  in \eqref{eq:conditional_f}. In that equation, the mean of the distribution of $f_p$ depends on both the precise value of the inducing points, $u_p$, and the kernel matrix, $K_{XZ}$. After conditioning on the observed data through the likelihood, $p(y_n| f_n)$, the posterior distribution for $U$ and $X$ becomes conditionally dependent. 

The main contribution of this paper is to allow the variational posterior to capture this dependence, by  relaxing the MF assumption explained in Equation~\ref{eq:MF-factorisation}. 
We do this by applying Structured Stochastic Variational Inference \citep{hoffman2015ssvi}, a general framework that relaxes the independence assumptions of MF inference. The insight of SSVI is to take the conditional independence structure of the full posterior and mirror it with a conditional independence structure of the variational posterior. Concretely, the variational posterior is assumed to have the factorization 
\begin{equation}
  q(F,X,U)= \prod_{p=1}^P p(f_{p}| u_p, X)  q(u_p) \prod_{n=1}^{N} q(x_n\mid U) .
  \label{eq:structured-factorisation}
\end{equation}
The key difference between this and Equation~\eqref{eq:MF-factorisation} is that now the variational posterior of the latent variables, $x_n$, depends on the global variables, $U$. This can be done while maintaining a parametric form of the posterior by keeping the same parametrization of the inducing points as the MF case where $q(u_p) = N(\mu_p, \Sigma_p)$, with $\mu_p$, $\Sigma_p$ being constant values, but now making the parameters of $q(x_n| U) = N(\mu_n, \Sigma_n)$ functions of the inducing points, \textit{i.e.} $
(\mu_n, \Sigma_n) = h(y_n, U)$ with $h: \R^{P}\times \R^{M\times P} \to \R^{Q}\times\R^{Q \times Q}$. The goal of the VI procedure is now to learn the constant values $\mu_p, \Sigma_p$, as well as the mapping $h(\cdot, \cdot)$ which most closely matches the true posterior. 



By once again noting that the KL contribution from the GP values, $F$, is zero due to the shared conditional distribution of $F$ given $U$ and $X$, the ELBO for the SSVI objective can be written as

\begin{equation}
\label{eq:SSVI_ELBO}
\mathcal L_S
=
\sum_{n=1}^N \mathbb E_{q(U)}\!\left[\hat{\mathcal L}_n(U)\right]
- D_{\mathrm{KL}}\!\left(q(U)\,\|\,p(U)\right)
\end{equation}
where
\begin{equation}
\label{eq:LnU}
\begin{aligned}
\hat{\mathcal L}_n(U)
&=
\mathbb E_{p(f_n\mid x_n,U)\,q(x_n\mid U)}
\!\left[\log p\!\left(y_n \mid f_n,x_n,U\right)\right]
- D_{\mathrm{KL}}\!\left(q(x_n\mid U)\,\|\,p(x_n)\right).
\end{aligned}
\end{equation}

The term within the expectation of the inducing points, $\hat \elbo_n(U)$\footnote{In order to fully define $\hat \elbo_n$ values for $y_n$, $\mu_n$, $\Sigma_n$, and parameters of the GP are all required in addition to $U$.  The inclusion of $U$ in this notation is to emphasize the dependence of $\hat \elbo_n(\cdot)$ on a particular value of the random variable, which is the key difference between the local ELBO and the standard ELBO.}, is called the Local ELBO. 
As in the Mean Field case, the SSVI ELBO decomposes into a term depending on the global variables $q(U)$, and a sum over terms which depend on the local variables $y_n$. This decomposition allows the objective to be stochastically optimized through mini-batching over the observations, allowing our method to scale to large datasets. 
The conditional variational distribution for the latent space, $q(x_n \mid U)$, may depend on $U$ in an arbitrarily complex way, making the expectation $\mathbb E_{q(U)}$ intractable. We therefore approximate it using $K$ Monte Carlo
samples $U_k \sim q(U)$, yielding
\begin{equation}
\label{eq:struc-elbo}
\mathcal L_S
\approx
\frac{1}{K}\sum_{k=1}^K \sum_{n=1}^{N} \hat{\mathcal L}_n(U_k)
- D_{\mathrm{KL}}\!\big(q(U)\,\|\,p(U)\big).
\end{equation}
This expression can be evaluated either using Monte Carlo techniques to evaluate the expectations, or, for special cases such as the ones we present in our examples, the expressions are analytically tractable. We present further details of our ELBO in Appendix~\ref{sec:elbo-derivation-details}.

\subsection{Local and Global Variables for Stochastic Variational Inference}
One important choice of our variational family which we have not discussed so far is the choice to specify our variational posterior as $q(X|U)q(U)$, rather than $q(U|X) q(X)$. The structure of the exact posterior can be written equally well in both forms: $p(X,U,F|Y) = p(F|U,X) p(U|X) p(X) = p(F|U,X) p(X|U) p(U)$. Instead, the reason to prefer $q(X|U)q(U)$ over the alternative is computational. 

We are interested in developing a method which can scale to a larger number of observations than can reasonably be expected to all be fit into memory at once. To do this, we need to use Stochastic Variational Inference \citep[SVI;][]{hoffman2013StochasticVariationalInference}. SVI relies on models whose variables can be split into two sets: local variables, which are associated with individual data points, and global variables, which are associated with many or all data points. The key insight of SVI is to realize that it is possible to get an unbiased estimate of the gradient for the parameters of the global variables by using only a subset of the data. Structured Stochastic Variational Inference \citep{hoffman2015ssvi} relies on this same insight, but with additional dependencies in the variational posterior. 

In GPs, $U$ are the global variables shared across the full dataset \citep{hensmanGaussianProcessesBig2013}, and in GPLVMs the local variables are latent variables, $x_n$. This means that the ELBO decomposes into a sum of per-data point contributions, which in turn allows us to estimate $\elbo_S$ using only an index set $\mathcal{I}$, where $|\mathcal{I}| \ll N$, as
\begin{equation}
\label{eq:struc-elbo-minibatch}
\tilde {\mathcal L}_S
= 
\frac{N}{K |\mathcal{I}|}\sum_{k=1}^K \sum_{n \in \mathcal{I}} \hat{\mathcal L}_n(U_k)
- D_{\mathrm{KL}}\!\big(q(U)\,\|\,p(U)\big).
\end{equation}
This mini-batched version of the objective is the one we use in our experiments.

\begin{table*}[t]
\centering
\caption{Performance comparison on the Oilflow, qPCR, MNIST, and FMNIST
datasets under the standard configuration. Values are reported as
mean~$\pm$~standard deviation over 5 runs. The final negative
ELBO is not reported for IWVI-GPLVM because it optimizes a different
importance-weighted objective that is not directly comparable to the standard
ELBO.}
\label{tab:oilflow-qpcr}
\resizebox{\textwidth}{!}{%
\begin{tabular}{lcccccc}
\toprule
Dataset & GPLVM type & Train RMSE & Test RMSE & Train NLPD & Test NLPD & Final neg. ELBO \\
\midrule
Oilflow & MF
& $0.0528 \pm 0.0040$ & $0.1581 \pm 0.0088$
& $-15.9416 \pm 0.5589$ & $28.5091 \pm 4.6286$
& $-5.2531 \pm 0.2447$ \\
& A-MF
& $0.0460 \pm 0.0045$ & $0.0496 \pm 0.0053$
& $-14.8375 \pm 0.5922$ & $-14.3680 \pm 0.9450$
& $-4.8859 \pm 0.2265$ \\
& IWVI-GPLVM
& $\mathbf{0.0343 \pm 0.0025}$ & $\mathbf{0.0375 \pm 0.0023}$
& $-17.8099 \pm 0.7425$ & $-17.4136 \pm 0.4695$
& -- \\
& A-S \textit{(Ours)}
& $0.0418 \pm 0.0032$ & $0.0479 \pm 0.0028$
& $\mathbf{-19.8808 \pm 0.7041}$ & $\mathbf{-18.8218 \pm 0.6210}$
& $\mathbf{-6.4527 \pm 0.0449}$ \\
\midrule
qPCR & MF
& $0.4993 \pm 0.0141$ & $0.6018 \pm 0.0405$
& $36.1396 \pm 1.0722$ & $50.4384 \pm 5.1887$
& $46.9602 \pm 0.2065$ \\
& A-MF
& $0.4795 \pm 0.0109$ & $0.5190 \pm 0.0061$
& $35.3542 \pm 0.6533$ & $38.7142 \pm 0.8894$
& $47.4945 \pm 0.2714$ \\
& IWVI-GPLVM
& $0.4354 \pm 0.0053$ & $0.4731 \pm 0.0175$
& $32.4221 \pm 0.4552$ & $35.5986 \pm 1.3582$
& -- \\
& A-S \textit{(Ours)}
& $\mathbf{0.4282 \pm 0.0047}$ & $\mathbf{0.4701 \pm 0.0132}$
& $\mathbf{29.8229 \pm 0.4723}$ & $\mathbf{33.5050 \pm 1.2980}$
& $\mathbf{44.9380 \pm 0.3163}$ \\
\midrule
MNIST & A-MF
& $0.1550 \pm 0.0006$ & $0.1568 \pm 0.0003$
& $-342.5288 \pm 2.8080$ & $-333.9984 \pm 1.5725$
& $-294.0345 \pm 2.8327$ \\
& IWVI-GPLVM
& $0.1540 \pm 0.0010$ & $0.1581 \pm 0.0011$
& $-348.0312 \pm 5.5166$ & $-328.9493 \pm 5.6709$
& -- \\
& A-S \textit{(Ours)}
& $\mathbf{0.1515 \pm 0.0001}$ & $\mathbf{0.1537 \pm 0.0007}$
& $\mathbf{-359.8706 \pm 2.1250}$ & $\mathbf{-348.4373 \pm 3.0311}$
& $\mathbf{-306.3421 \pm 1.0179}$ \\
\midrule
FMNIST & A-MF
& $0.1339 \pm 0.0004$ & $0.1342 \pm 0.0006$
& $-456.6282 \pm 3.1032$ & $-454.4614 \pm 4.2266$
& $-409.8799 \pm 2.5766$ \\
& IWVI-GPLVM
& $0.1340 \pm 0.0009$ & $0.1380 \pm 0.0011$
& $-458.0779 \pm 5.4105$ & $-436.8031 \pm 6.0015$
& -- \\
& A-S \textit{(Ours)}
& $\mathbf{0.1294 \pm 0.0003}$ & $\mathbf{0.1299 \pm 0.0004}$
& $\mathbf{-484.1356 \pm 1.8203}$ & $\mathbf{-481.3409 \pm 2.7219}$
& $\mathbf{-425.9231 \pm 1.0856}$ \\
\bottomrule
\end{tabular}%
}
\end{table*}
\label{sec:amortized-encoder}

\begin{figure*}[t]
    \centering

    \begin{subfigure}[t]{0.23\linewidth}
        \centering
        \includegraphics[width=\linewidth]{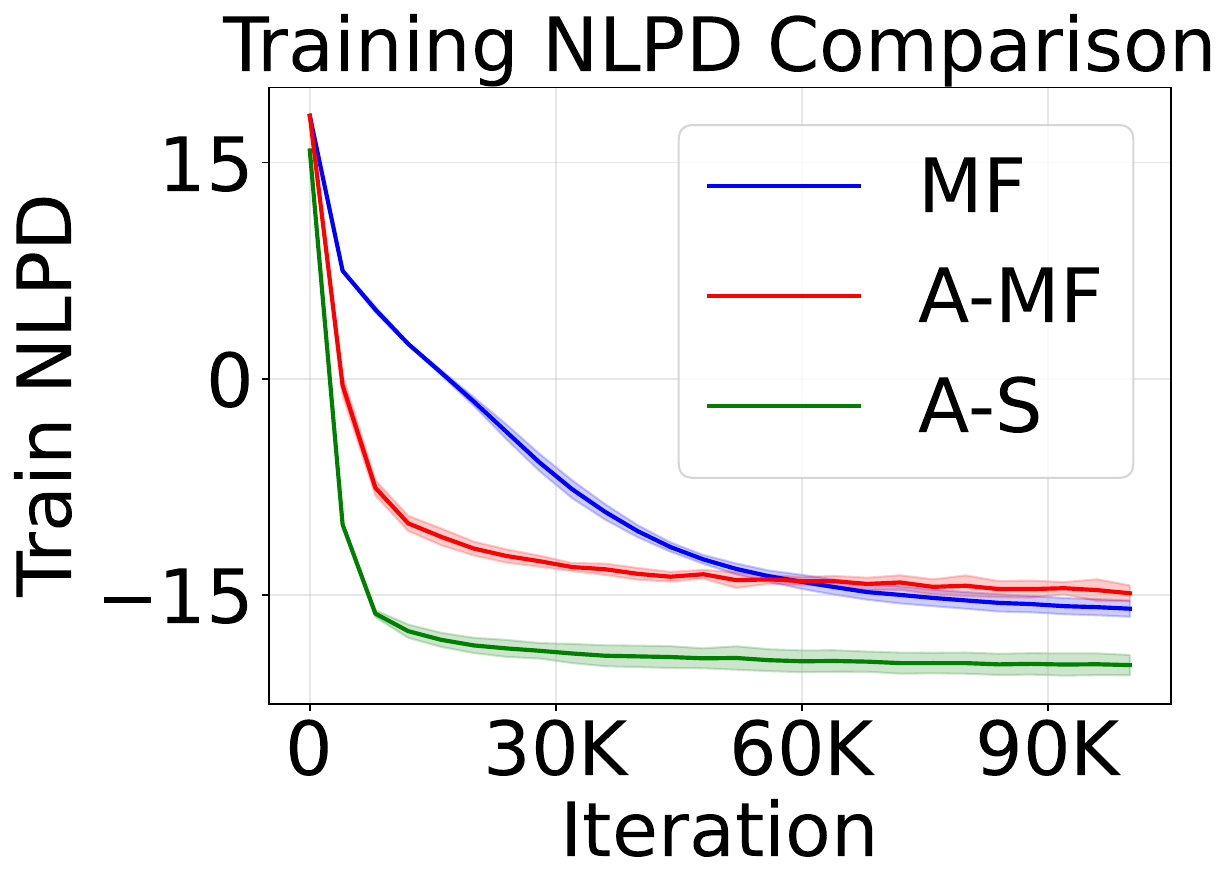}
        \caption{Oilflow }
        \label{fig:nlpd-oilflow}
    \end{subfigure}
    \begin{subfigure}[t]{0.23\linewidth}
        \centering
        \includegraphics[width=\linewidth]{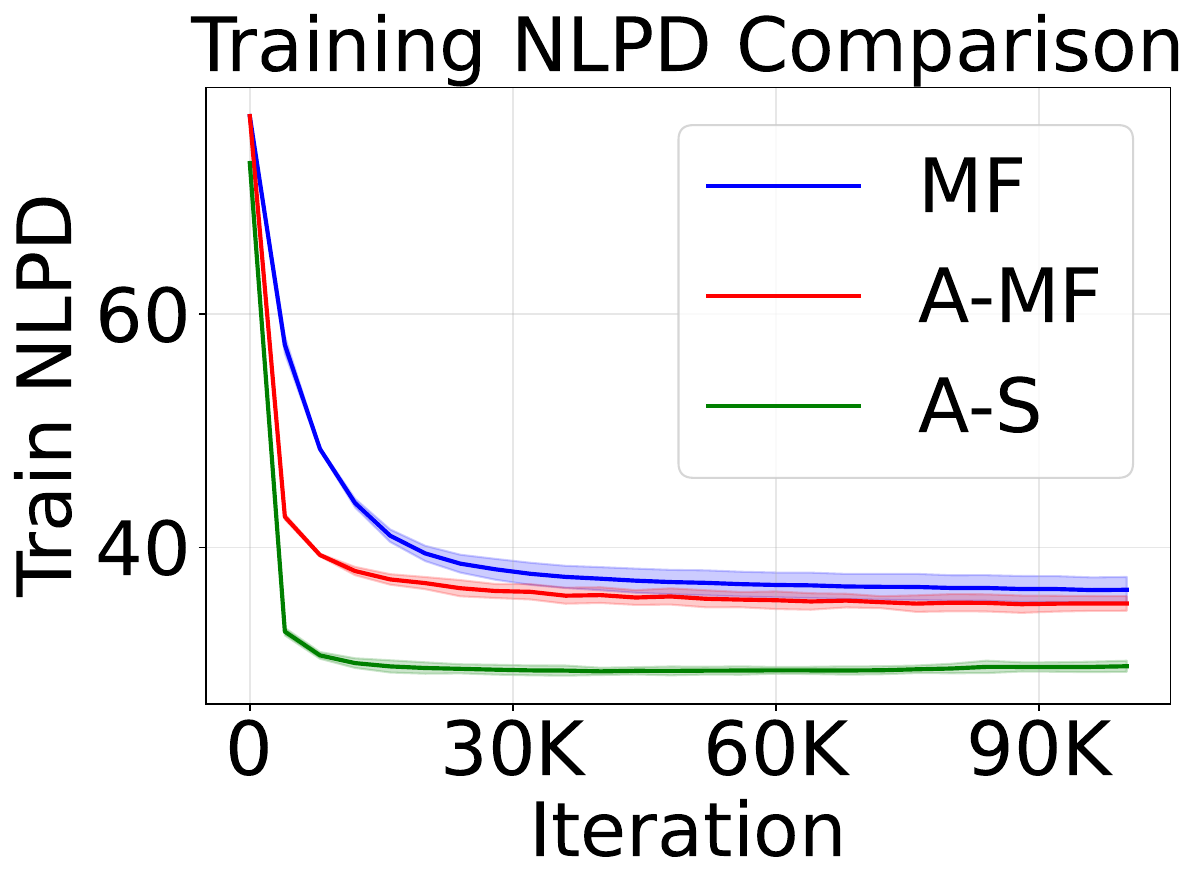}
        \caption{qPCR }
        \label{fig:nlpd-qpcr}
    \end{subfigure}
    \begin{subfigure}[t]{0.23\linewidth}
        \centering
        \includegraphics[width=\linewidth]{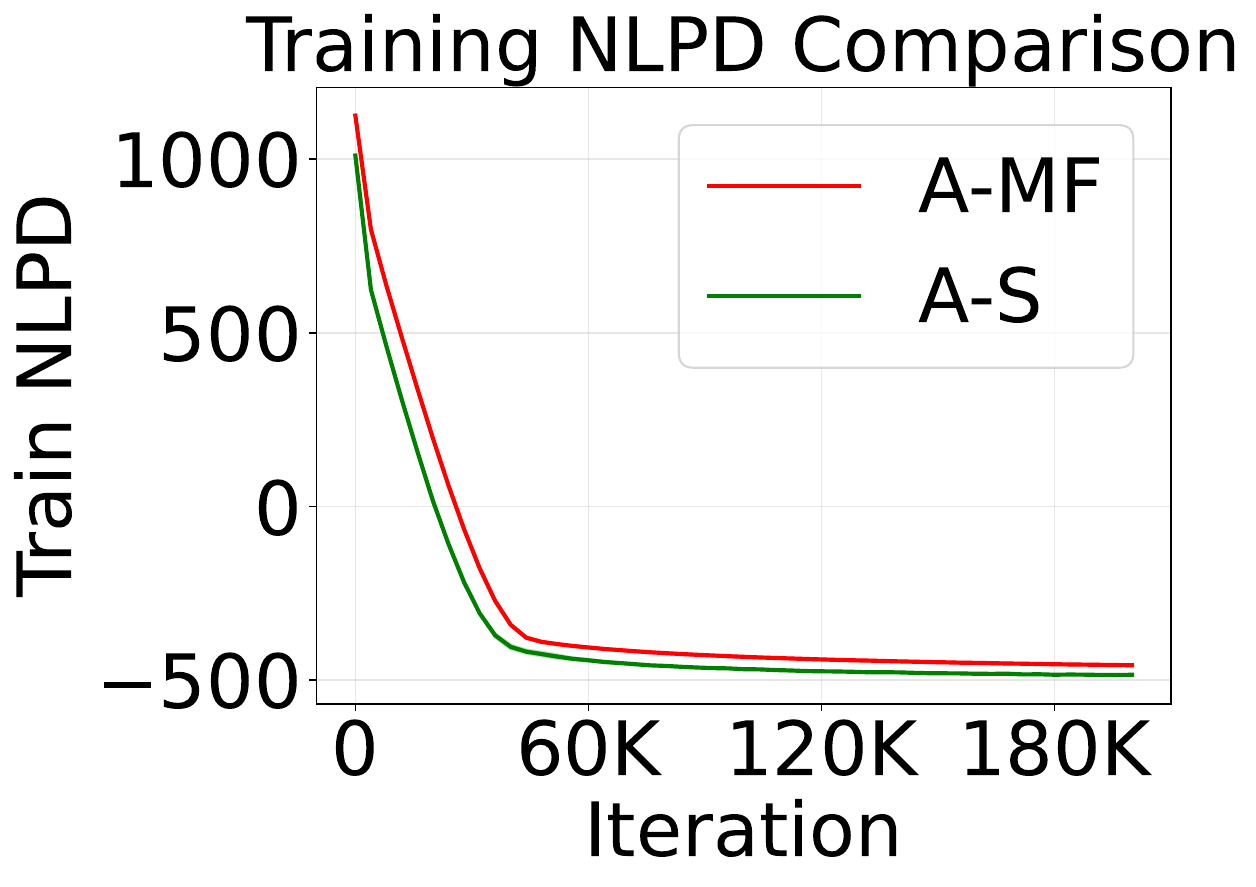}
        \caption{FMNIST }
        \label{fig:nlpd-fmnist}
    \end{subfigure}
    \begin{subfigure}[t]{0.23\linewidth}
        \centering
        \includegraphics[width=\linewidth]{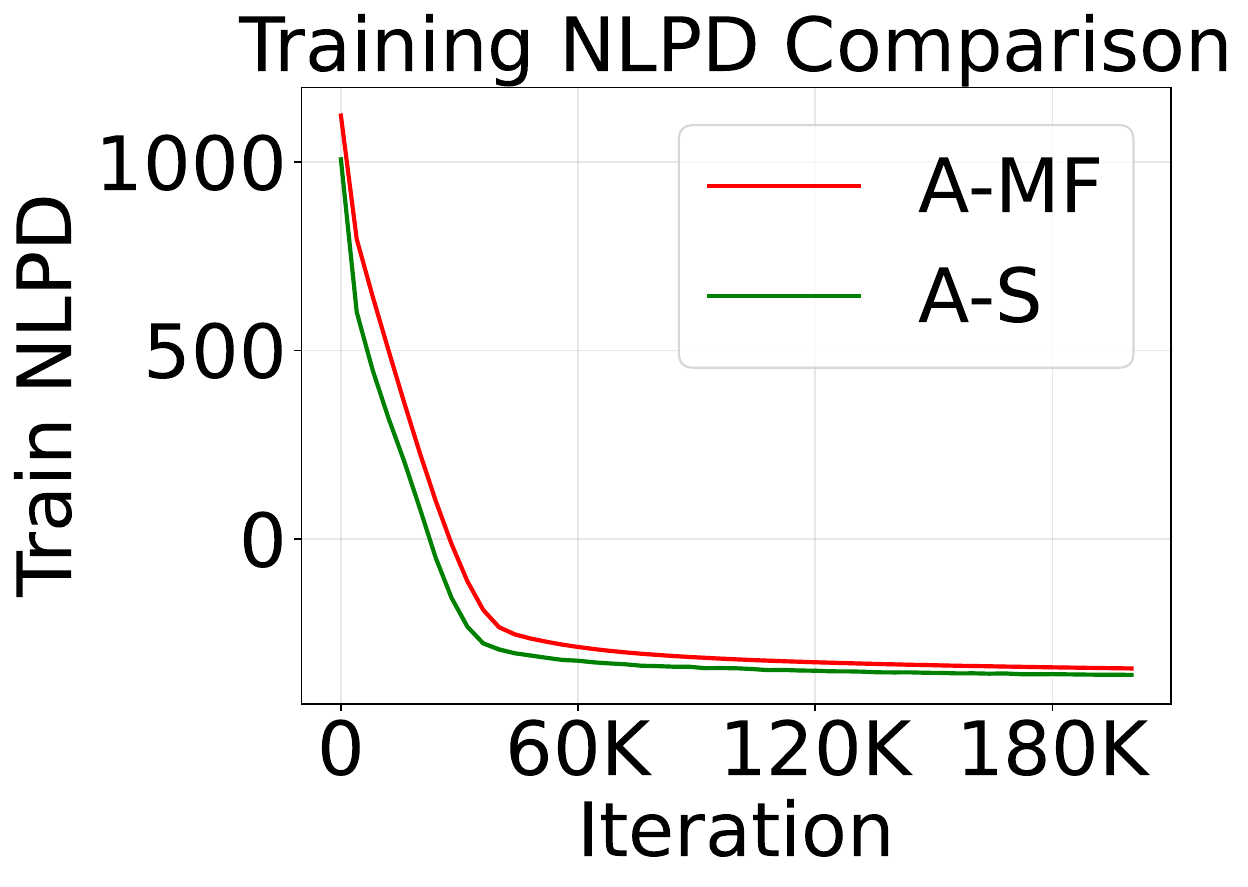}
        \caption{MNIST }
        \label{fig:nlpd-mnist}
    \end{subfigure}
    \caption{The training NLPD progression during training across all datasets. The A-S-GPLVM and  A-MF-GPLVM converge at similar rates, but the A-S-GPLVM converges to a better optimum.}
    \label{fig:nlpd-all}
\end{figure*}

\begin{figure}[t]
    \centering

    \begin{subfigure}[t]{0.3\linewidth}
        \centering
        \includegraphics[width=\linewidth]{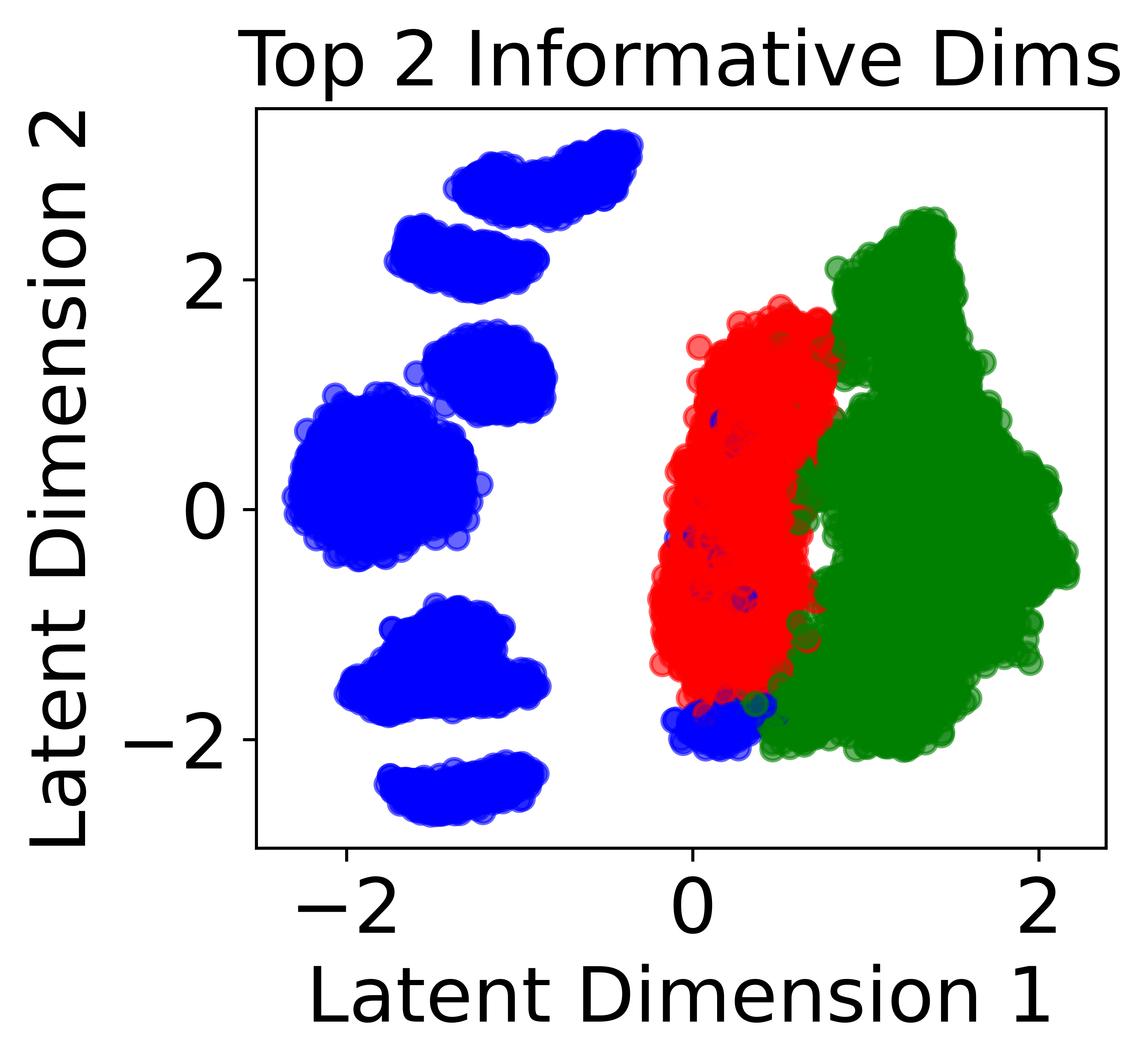}
        \caption{MF-GPLVM}
        \label{fig:latent-svi}
    \end{subfigure}
    \begin{subfigure}[t]{0.3\linewidth}
        \centering
        \includegraphics[width=\linewidth]{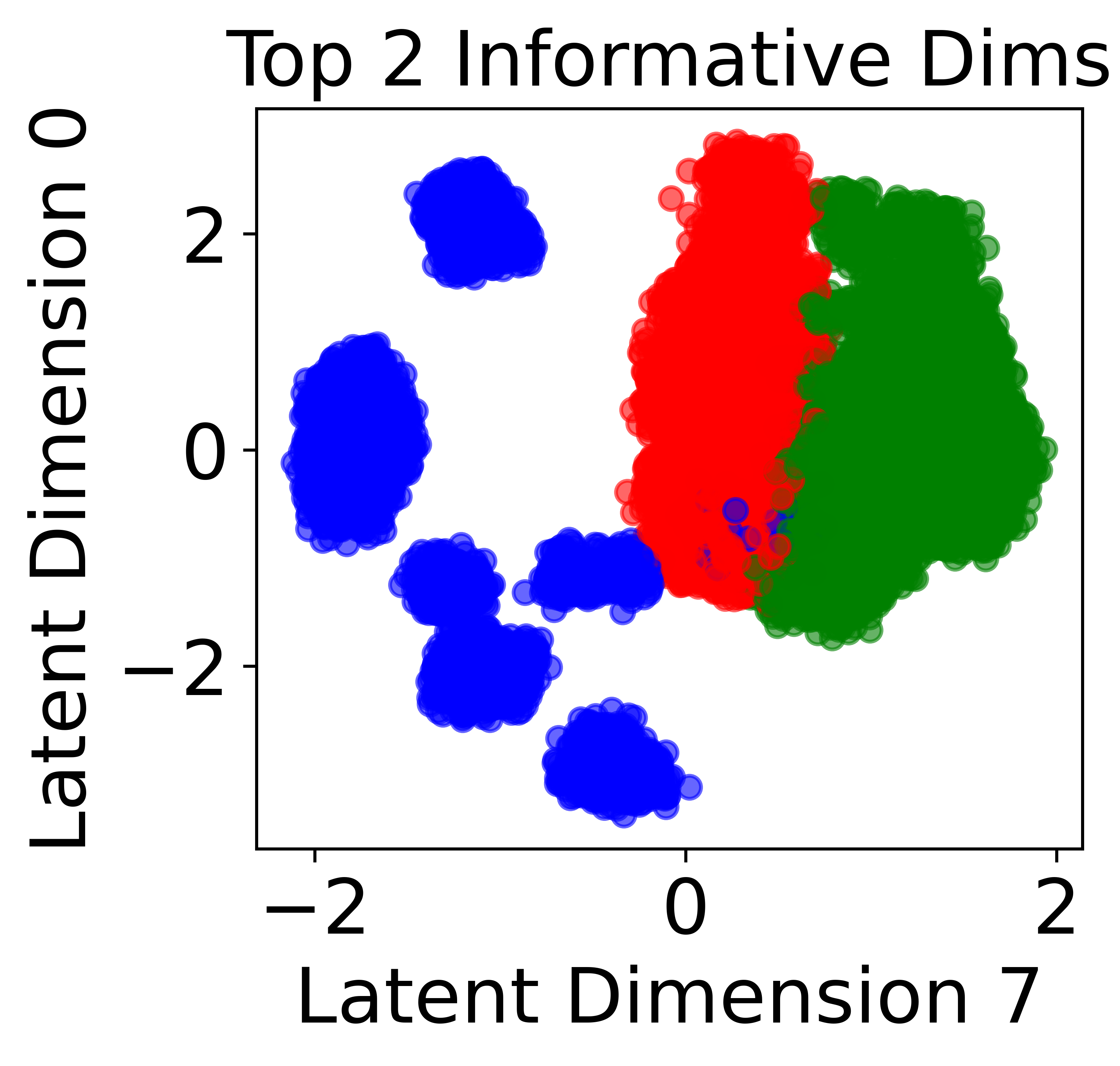}
        \caption{A-MF-GPLVM}
        \label{fig:latent-amf}
    \end{subfigure}
    \begin{subfigure}[t]{0.3\linewidth}
        \centering
        \includegraphics[width=\linewidth]{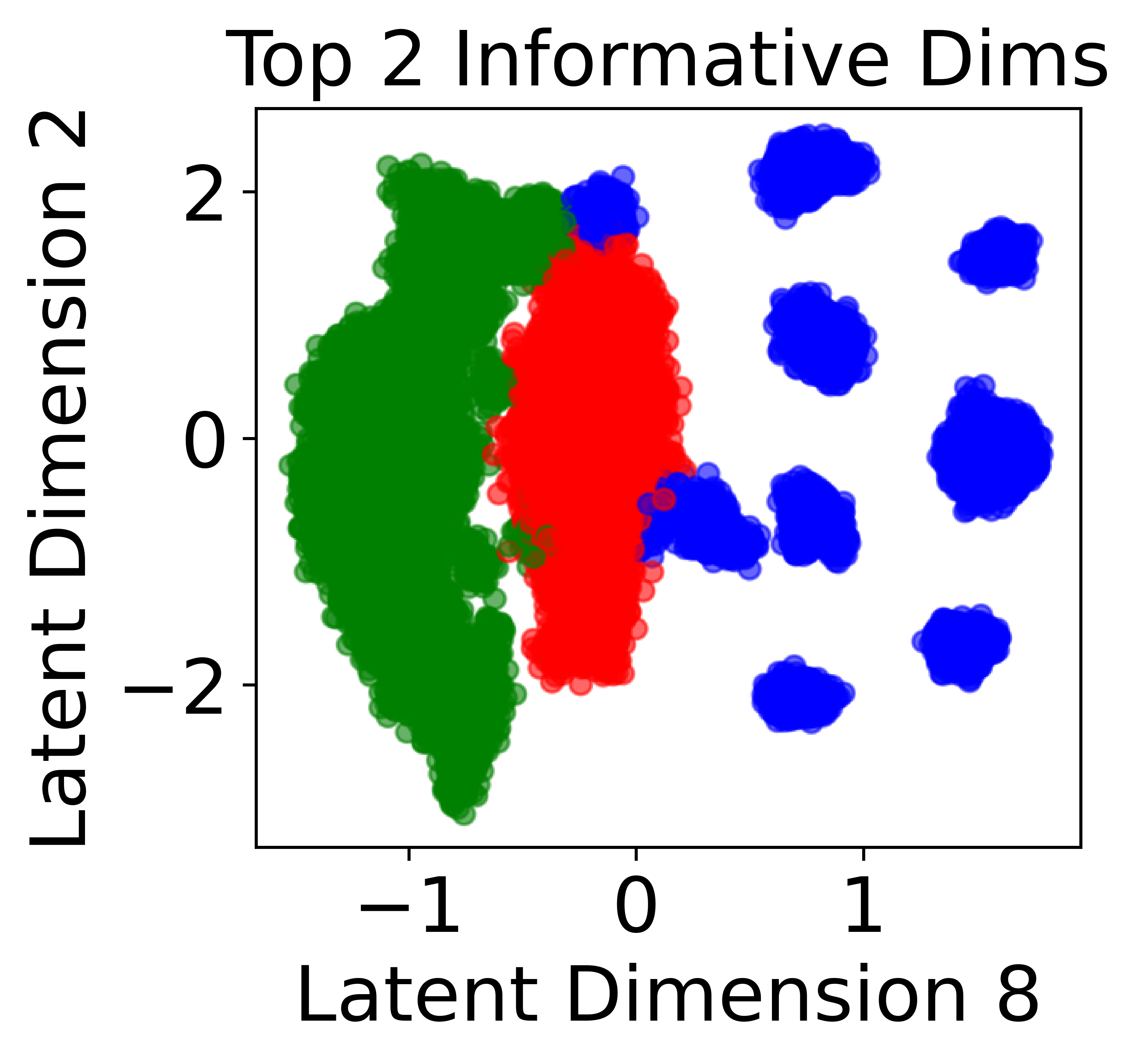}
        \caption{A-S-GPLVM}
        \label{fig:latent-as}
    \end{subfigure}
    \caption{
    The three latent spaces and reciprocal length scales for the oil flow dataset. The A-S-GPLVM has found a slightly more tightly clustered latent space and has been more aggressive about turning off non-informative dimensions.
    }
    \label{fig:standard-results}
\end{figure}

\begin{figure}[t]
\centering

\begin{subfigure}[t]{0.4\linewidth}
    \centering
    \includegraphics[width=\linewidth]{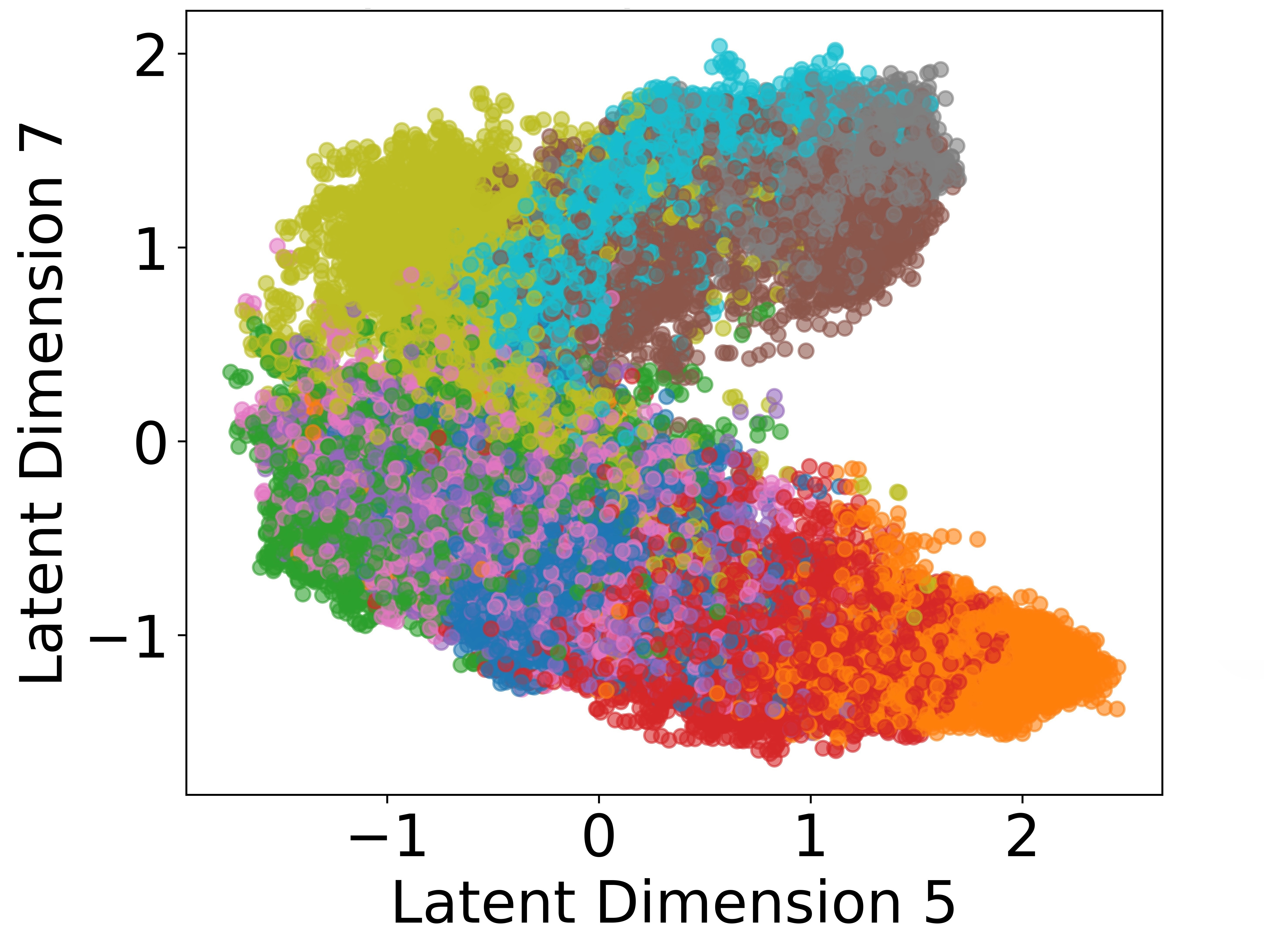}
    \subcaption{A-S-GPLVM.}
    \label{fig:fmnist-ssvi-latent-plot}
\end{subfigure}
\hspace{0.02\linewidth}
\begin{subfigure}[t]{0.4\linewidth}
    \centering
    \includegraphics[width=\linewidth]{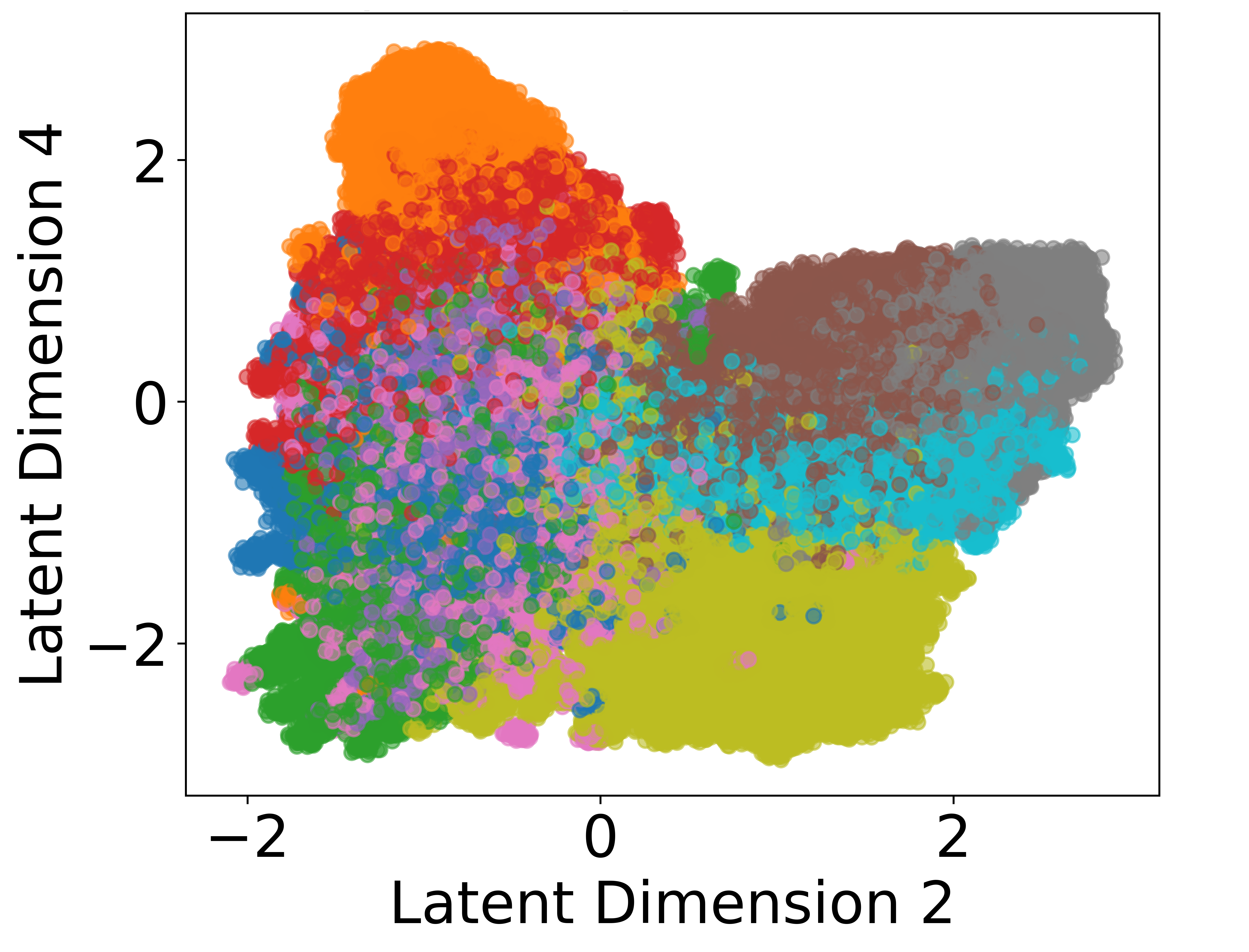}
    \subcaption{A-MF-GPLVM.}
    \label{fig:fmnist-a-mf-latent-plot}
\end{subfigure}

\vspace{0.5em}

\includegraphics[width=0.75\linewidth]{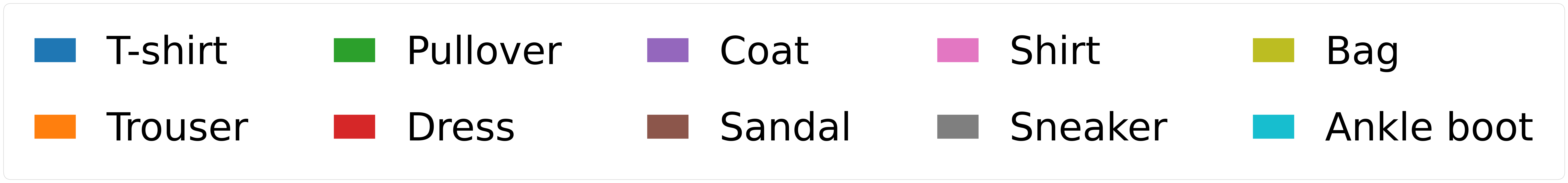}

\vspace{0.3em}

\caption{The distribution of training points in latent space for FMNIST for the A-S-GPLVM and A-MF-GPLVM. The A-S-GPLVM latent space has learned substantially more structure than the A-MF-GPLVM. In fact, the A-S-GPLVM appears to have collapsed the latent predictions into a semantically meaningful curve within the latent space, capturing the relative similarity of the classes with sandals, sneakers, and ankle boots at one end and trousers and dresses at the other. The A-MF-GPLVM seems to try to capture a similar shape, but achieves a much less clear separation between the groupings.}
\label{fig:fmnist-latent-plots}

\end{figure}

\begin{table*}[t]
\centering
\caption{Predictive uncertainty evaluation on the Oilflow, qPCR, MNIST, and
FMNIST datasets. Values are reported as mean~$\pm$~standard deviation over
5 runs. The best result
for each dataset and metric is shown in bold.}
\label{tab:crps-coverage}
\resizebox{\textwidth}{!}{%
\begin{tabular}{lccccc}
\toprule
Dataset & GPLVM type
& Train CRPS $\downarrow$
& Test CRPS $\downarrow$
& Train coverage
& Test coverage \\
\midrule
Oilflow & MF
& $0.0423 \pm 0.0031$
& $0.0832 \pm 0.0114$
& $0.9867 \pm 0.0026$
& $\mathbf{0.9385 \pm 0.0191}$ \\
& A-MF
& $0.0400 \pm 0.0020$
& $0.0416 \pm 0.0030$
& $0.9912 \pm 0.0045$
& $0.9862 \pm 0.0074$ \\
& IWVI-GPLVM
& $\mathbf{0.0238 \pm 0.0020}$
& $\mathbf{0.0251 \pm 0.0015}$
& $0.9979 \pm 0.0006$
& $0.9954 \pm 0.0026$ \\
& A-S \textit{(Ours)}
& $0.0247 \pm 0.0021$
& $0.0274 \pm 0.0021$
& $\mathbf{0.9863 \pm 0.0019}$
& $0.9705 \pm 0.0097$ \\
\midrule
qPCR & MF
& $0.3454 \pm 0.0061$
& $0.4077 \pm 0.0165$
& $\mathbf{0.9587 \pm 0.0024}$
& $0.9383 \pm 0.0116$ \\
& A-MF
& $0.3364 \pm 0.0031$
& $0.3573 \pm 0.0089$
& $0.9650 \pm 0.0020$
& $\mathbf{0.9477 \pm 0.0076}$ \\
& IWVI-GPLVM
& $0.2459 \pm 0.0030$
& $0.2613 \pm 0.0049$
& $0.9763 \pm 0.0005$
& $0.9689 \pm 0.0047$ \\
& A-S \textit{(Ours)}
& $\mathbf{0.2381 \pm 0.0015}$
& $\mathbf{0.2558 \pm 0.0079}$
& $0.9686 \pm 0.0007$
& $0.9589 \pm 0.0056$ \\
\midrule
MNIST & A-MF
& $0.1007 \pm 0.0001$
& $0.1016 \pm 0.0001$
& $0.9269 \pm 0.0001$
& $0.9247 \pm 0.0004$ \\
& IWVI-GPLVM
& $0.1008 \pm 0.0009$
& $0.1024 \pm 0.0010$
& $\mathbf{0.9314 \pm 0.0008}$
& $\mathbf{0.9272 \pm 0.0008}$ \\
& A-S \textit{(Ours)}
& $\mathbf{0.0978 \pm 0.0001}$
& $\mathbf{0.0989 \pm 0.0002}$
& $0.9270 \pm 0.0002$
& $0.9248 \pm 0.0004$ \\
\midrule
FMNIST & A-MF
& $0.0889 \pm 0.0001$
& $0.0890 \pm 0.0002$
& $0.9330 \pm 0.0001$
& $0.9324 \pm 0.0004$ \\
& IWVI-GPLVM
& $0.0898 \pm 0.0008$
& $0.0916 \pm 0.0005$
& $\mathbf{0.9371 \pm 0.0008}$
& $0.9321 \pm 0.0027$ \\
& A-S \textit{(Ours)}
& $\mathbf{0.0859 \pm 0.0003}$
& $\mathbf{0.0861 \pm 0.0001}$
& $0.9337 \pm 0.0002$
& $\mathbf{0.9333 \pm 0.0007}$ \\
\bottomrule
\end{tabular}%
}
\end{table*}

\section{Related work}
\label{sec:related-works}

A closely related work is the GP state-space model work of \citet{ialongo2019overcoming}, who remove the mean-field factorization between latent state trajectories and the GP transition posterior in recurrent GP models. Their coupled approximation improves calibration while preserving similar time and space complexity as the mean-field alternative. This work approaches the same conceptual challenge: that decoupling latent variables from the GP posterior can be too restrictive, but \citet{ialongo2019overcoming} study sequential GP state-space models, whereas we consider the latent variable setting.

A complementary approach to improving variational inference in GP models is to tighten the variational objective through importance weighting. \citet{salimbeni2019deep} develop this approach for deep Gaussian processes, providing a mechanism to trade additional computation for a tighter lower bound. Adapted to the sparse GPLVM setting, importance-weighted variational inference offers an alternative to our explicit coupling of the latent-variable and inducing-variable posteriors. An amortized IWVI-GPLVM therefore provides a relevant baseline for comparing these two approaches to improving variational inference while retaining the same underlying sparse GPLVM model.

Besides the Bayesian GPLVM \citep{titsias2010bayesian} and Generalised GPLVM \citep{lalchandGeneralisedGaussianProcess2022}, discussed in the background section,  there have been several other lines of work within the GPLVM field. 
One such line of work is the Random Feature Latent Variable Model (RFLVM), which replaces the sparse variational GP approximation with Random Fourier Feature (RFF) approximation of the mapping from latent variables to observations \citep{gundersen2021latent}.  In their approach \citet{gundersen2021latent} used a point MAP estimate for the latent variables $X$, which was conditioned on the RFF approximation. Later work in this line by \citet{li2024preventing} and \citet{yang2025multi} applied a VI approach to approximating the RFLVM posterior. A different approach by \citet{xu2024variational} used the same sparse GP formulation of the mapping from the latent to observed space, but used Annealed Importance Sampling (AIS) to tighten the lower bound of the GPLVM. 
As far as we are aware, no prior work has proposed a method for making the latent space conditional on the GP mapping as we do here.

\section{Experiments}
\label{sec:experiments}


\begin{figure}[t]
\centering
\begin{subfigure}[t]{0.48\linewidth}
    \centering
    \includegraphics[width=\linewidth]{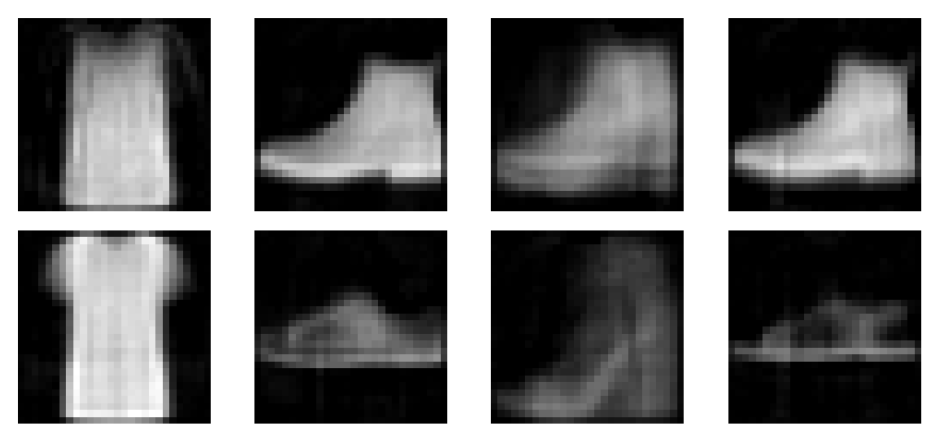}
    \subcaption{FMNIST }
    \label{fig:fmnist-ssvi-reconstruction}
\end{subfigure}\hfill
\begin{subfigure}[t]{0.48\linewidth}
    \centering
    \includegraphics[width=\linewidth]{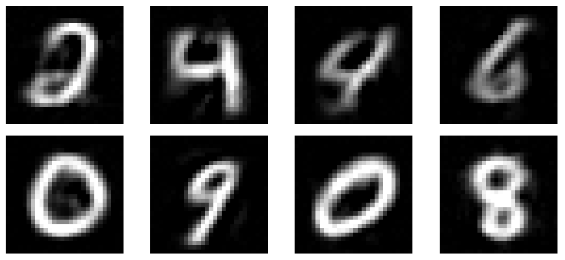}
    \subcaption{MNIST}
    \label{fig:mnist-ssvi-reconstruction}
\end{subfigure}
\caption{Sample reconstructions using the A-S-GPLVM. Each image shows the posterior predictive mean of the reconstructed observation. The model is able to produce a reasonable quality of reconstruction, although the difference in quality to the A-MF-GPLVM (see Figure~\ref{fig:mf-reconstructions}, Appendix~\ref{sec:additional-plots}) is not immediately obvious. }
\label{fig:ssvi-reconstructions}
\end{figure}

We compare our extension of the GPLVM framework, the Amortized-Structured-GPLVM (A-S-GPLVM), which applies SSVI with amortization of the latent variables, to variants of the mean field approach. On the smaller scale datasets we compare against two inference strategies: the MF-GPLVM with SVI, which directly approximates the MF-GPLVM variational posterior by optimizing the parameters of the latent variable distribution $\{\mu_n, \Sigma_n\}_{n=1}^N$ and the inducing point variational distribution $\{\mu_p, \Sigma_p\}_{p=1}^P$, and the  Amortized-MF-GPLVM (A-MF-GPLVM), where the latent variables are also amortized. We base our implementation of these on \citet{lalchandGeneralisedGaussianProcess2022}. The initialization is the same for all models, and is reported in Appendix~\ref{sec:model-config}. A Bayesian diagram for the A-MF-GPLVM and A-S-GPLVM is given in Figure~\ref{fig:gplvm-three}, in Appendix~\ref{sec:model-config}. For larger datasets, we do not include the non-amortized MF-GPLVM.

Motivated by the alternative bound-tightening approach discussed in Section~\ref{sec:related-works}, we additionally compare against an amortized importance-weighted variational GPLVM (IWVI-GPLVM), following the importance-weighted approach of \citet{salimbeni2019deep}. This baseline retains the same underlying sparse GPLVM model but improves the variational objective through importance weighting rather than introducing the explicit posterior dependence used by A-S-GPLVM. It therefore allows us to assess whether the benefits of structured inference persist when compared with an alternative approach to tightening the variational bound. We use the same model and optimization hyperparameters as for the other amortized models, with the additional IWVI-specific parameter, the number of importance samples, set to $5$.

\paragraph{Data setup.} 
The experiments are conducted on four standard datasets: two small-scale ones and two larger datasets. Our small-scale datasets are the Oilflow dataset, consisting of 1000 twelve-dimensional data points belonging to three classes corresponding to different phases of oil flow in a pipeline \citep{bishop1993analysis}, and the qPCR dataset, consisting of 48-dimensional single-cell data obtained from mice with 10 classes \citep{guo2010resolution}. For our larger-scale experiments, we use the MNIST \citep{lecun1998gradient} and Fashion MNIST \citep{xiao2017fashion}, which we subsample to contain $1500$ examples from each of the 10 classes. Code to recreate the Oilflow experiment using the A-S-GPLVM model is provided in the accompanying repository (see \url{https://github.com/jamesacodgers/conditional_overlaping_gaussian_processes}).

\paragraph{Evaluation Metrics.} 
To evaluate the quality of our posterior approximation, we use five metrics. First, we use the negative ELBO from our model, which should give a lower value if our posterior is able to achieve a value closer to the true posterior. As the ELBO is also influenced by hyperparameters, which can differ between models, we also use additional metrics to evaluate the reconstruction quality of our learned data manifold. We define our variational reconstruction distribution with 
\begin{equation}
q(y_n) = \int p(y_n|f_n) q(f_n, x_n, U) d f_n dx_n dU,
\end{equation}
for each of our data points. We report two values calculated from this: the Negative Log Predictive Density (NLPD) of $q(y_n)$ evaluated at the true data point, and the Root Mean Squared Error (RMSE) between the true observation $y_n$ and the mean of $q(y_n)$. 

To evaluate predictive uncertainty more directly, we additionally report the Continuous Ranked Probability Score (CRPS) and empirical coverage of marginal $95\%$ posterior predictive intervals. CRPS evaluates the full predictive distribution and rewards both calibration and sharpness, with lower values indicating better predictions. Empirical coverage measures the fraction of observed values contained in the corresponding $95\%$ predictive intervals; values closer to $0.95$ indicate better calibration.

We also split our data into training and test sets. The training set is the data used to train the GP hyperparameters, the inducing point parameters, and either the values of the latent variables or the amortization network for the latent variables. The test data were not seen during training, and we use this data to evaluate if our model can generalize well to unseen observations. For the smaller datasets, we use a 90/10 train–test split, while for the MNIST and FMNIST datasets, we used 1000 training points and 500 test points.

\paragraph{Data pre-processing.}
For all datasets, we apply mean-centering as a preprocessing step.
Specifically, each observed dimension is centered by subtracting its empirical mean, computed over the training set.

\subsection{Results.}

\begin{figure}[t!]
  \centering
  \subcaptionbox{MF-GPLVM\label{fig:latent-slice-gauss}}{%
    \includegraphics[width=0.32\linewidth]{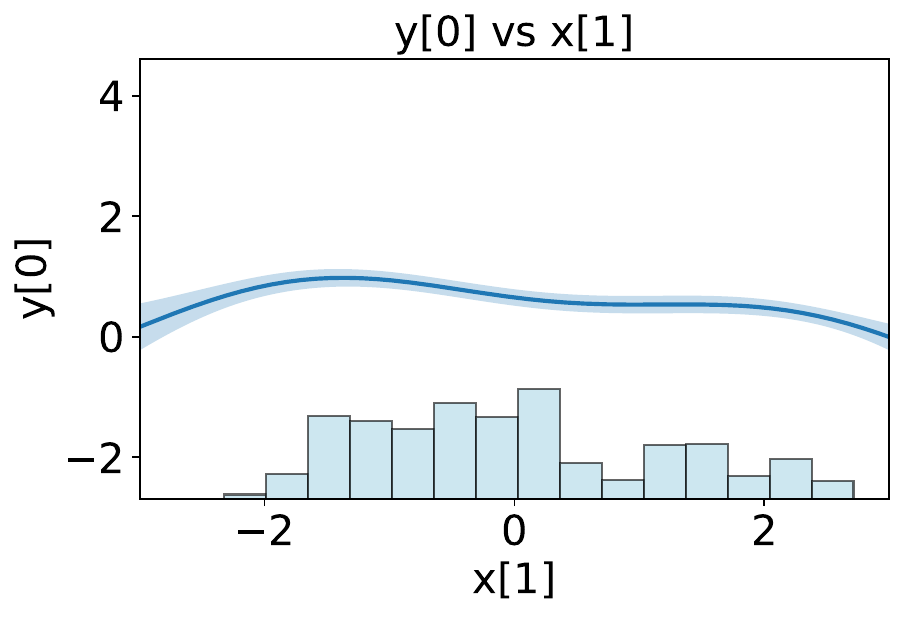}
  }
  \subcaptionbox{A-MF-GPLVM\label{fig:latent-slice-nn}}{%
    \includegraphics[width=0.32\linewidth]{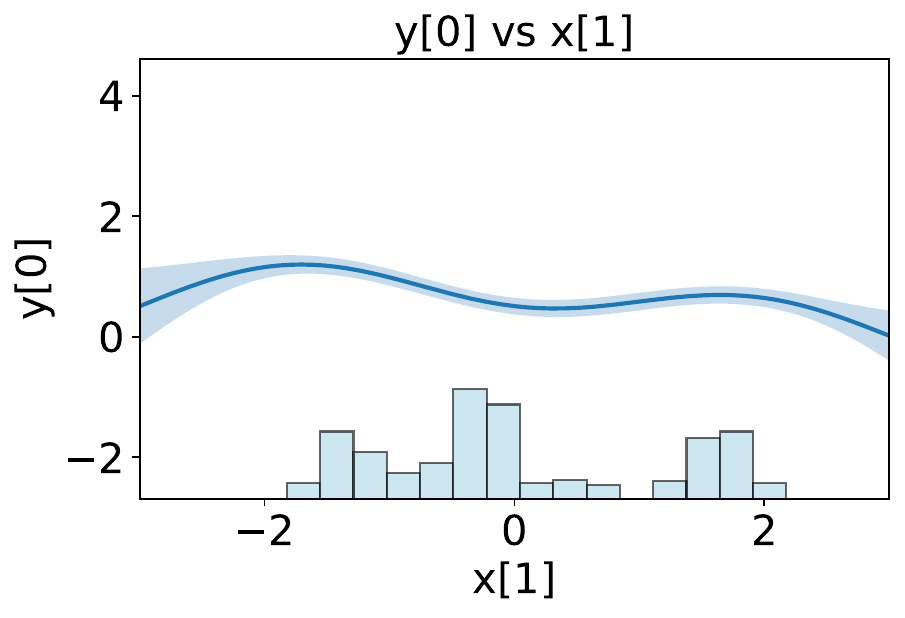}
  }
  \subcaptionbox{A-S-GPLVM\label{fig:latent-slice-ssvi}}{%
    \includegraphics[width=0.32\linewidth]{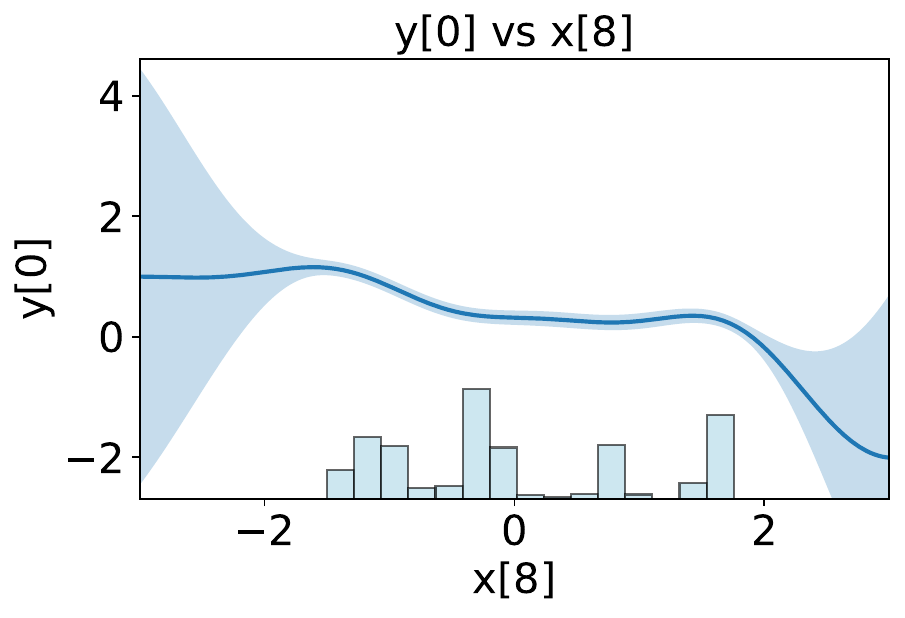}
  }
  \caption{Latent slice through the most informative latent dimension and a histogram of the learned posterior distribution of the training data along the horizontal axis. 
  Curves show predictive means; shaded areas denote predictive uncertainty for each output. The epistemic uncertainty at points far from the training data clearly increases more rapidly with the A-S-GPLVM than with the MF models. 
  }
  \label{fig:latent-slice}
\end{figure}

Table~\ref{tab:oilflow-qpcr} reports the train and test RMSE and NLPD, as well as the final negative ELBO where applicable. Compared with the A-MF-GPLVM, the A-S-GPLVM achieves lower RMSE and NLPD on all four datasets. Two-sided Welch's $t$-tests show that these improvements are statistically significant, with the exception of the train and test RMSE improvements on Oilflow.

The comparison with IWVI-GPLVM is more mixed on the smaller datasets. IWVI-GPLVM achieves lower train and test RMSE on Oilflow, whereas the A-S-GPLVM achieves significantly lower train and test NLPD. On qPCR, the A-S-GPLVM achieves significantly lower train and test NLPD, while its lower RMSE is not statistically significant. On MNIST and FMNIST, the A-S-GPLVM achieves significantly lower train and test RMSE and NLPD than IWVI-GPLVM.

Table~\ref{tab:crps-coverage} provides a more direct evaluation of predictive uncertainty. The A-S-GPLVM achieves significantly lower train and test CRPS than the A-MF-GPLVM on all four datasets. Compared with IWVI-GPLVM, the A-S-GPLVM achieves significantly lower train and test CRPS on MNIST and FMNIST. On Oilflow, IWVI-GPLVM obtains slightly lower CRPS, but the differences are not statistically significant. On qPCR, the A-S-GPLVM obtains lower CRPS, with a statistically significant difference on the training set. The empirical coverage results are more mixed and do not show uniform dominance by any single method. Nevertheless, the A-S-GPLVM achieves coverage closer to the nominal $95\%$ level than IWVI-GPLVM on both Oilflow and qPCR.

The poor performance of the non-amortized MF-GPLVM on test data is because $q(x_n)$ needs to be numerically optimized for each test point, which is a non-convex optimization and can converge to poor local optima. Figure~\ref{fig:nlpd-all} shows the convergence of the training NLPD across the four datasets. The A-S-GPLVM seems to converge at very similar rates to the A-MF-GPLVM, but to a better final value. In Appendix~\ref{sec:additional-plots}, Figures~\ref{fig:qpcr-metrics}-\ref{fig:mnist-metrics} show similar plots for the other metrics we tested.
The strong performance of the A-S-GPLVM compared to the MF variants suggests that our approach is using the greater flexibility of the variational family to achieve a better approximation of the exact posterior, and hence better performance.

We show two-dimensional projections of the latent space onto the two most significant dimensions of the GP, as in \citet{titsias2010bayesian}, for the oil flow and FMNIST datasets in Figure~\ref{fig:standard-results} and Figure~\ref{fig:fmnist-latent-plots}. 
For A-S-GPLVM, latent samples are obtained by first sampling the inducing variables $U$ from their variational posterior $q(U)$, finding $q(X \mid U)$, and sampling from that latent distribution. For the MF-GPLVM and the A-MF-GPLVM, latent samples are generated by sampling $q(X)$ directly. In both figures, the GPLVM variants all find comparable latent spaces, but the A-S-GPLVM produces results with tighter clusters than the (A-)MF-GPLVM. This is particularly pronounced for the FMNIST dataset in Figure~\ref{fig:fmnist-latent-plots}, where the A-S-GPLVM latent space has taken on a crescent shape, with one end containing footwear (sandal, sneaker, ankle boot), the other end taking clothing items (trouser, dress), and a clear gap containing no data between them. The A-MF-GPLVM also seems to be attempting to capture a similar shape; however, the separation between the two prongs is much less pronounced. This shows that, for this dataset, our method more effectively captures meaningful semantic information from the images. Figure~\ref{fig:standard-results} also plots the reciprocal of the ARD kernel's learned length scales. The A-S-GPLVM has been more aggressive in switching off uninformative dimensions, but learned comparable length scales for the informative dimensions. 
The latent plots for the qPCR and MNIST datasets are shown in
Appendix~\ref{sec:additional-plots} and exhibit similar behavior.


Figure~\ref{fig:ssvi-reconstructions} shows the mean of $q(y_n)$ from the A-S-GPLVM for eight data points, demonstrating reasonable reconstruction quality.  We provide similar plots for the MF-GPLVM reconstructions in Figure~\ref{fig:mf-reconstructions}, Appendix~\ref{sec:additional-plots}, which are visually of similar quality. While the improvements of the A-S-GPLVM reconstructions over the A-MF-GPLVM are mostly statistically significant, as shown in Table~\ref{tab:significance_tests}, the visual similarity suggests that further work improving other aspects of the GPLVM model would be needed to make it a truly effective generative model.

While the reconstructions are an important measure of latent space estimation quality, a specific advantage of GPs is their ability to increase epistemic uncertainty away from the data. We visualize this in Figure~\ref{fig:latent-slice}, which shows a slice of the prediction of the GP from the latent to one of the observed dimensions along the single most informative latent coordinate, \textit{i.e.}, the coordinate with the shortest GP length scale, with other coordinates fixed at zero. Figure~\ref{fig:latent-slice} shows that the MF-GPLVMs have a slow increase in the epistemic uncertainty of the GP mapping between the region where there is a significant probability of observing a latent variable, roughly corresponding to $[-2,2]$ on the $x$~axis, and the region where there is a low probability of observing a latent variable. In contrast, the uncertainty of the GP mapping learned by the A-S-GPLVM
increases much more rapidly away from the high-probability inputs.  We show similar plots for four output dimensions of the Oilflow dataset in Figure~\ref{fig:latent-slice-oilflow-app}, Appendix~\ref{sec:additional-plots}.  We do not expect the A-S-GPLVM to achieve a perfect estimation of the exact epistemic uncertainty of the posterior due to the approximation on the GP mapping \citep{wenger2022posterior, wenger2024computation} and other restrictions of the variational posterior. However, the A-S-GPLVM shows a clear improvement in the epistemic uncertainty estimation and,  as all of the methods presented here share the same GP approximation, these results indicate that the coupling of the variational posteriors for the GP and the latent space is directly responsible for this difference in epistemic uncertainty estimation. 


\section{Limitations}
\label{sec:limitations}

\paragraph{Computational overhead.}
Compared with the amortized mean-field GPLVM, the A-S-GPLVM introduces additional computational cost because the structured objective requires Monte Carlo samples $U_k \sim q(U)$, as in Equation~\ref{eq:struc-elbo}. For each sampled global state, the encoder must evaluate a conditional local posterior $q(x_n \mid U_k)$ and the expected likelihood must be evaluated. The $K$ sampled inducing-variable states can be processed in parallel, so the dominant practical cost is often increased memory usage rather than a fully sequential increase in wall-clock time.
We empirically quantify this in Appendix~\ref{sec:runtime-memory}, where we show that the A-S-GPLVM suffers a slight wall-clock slowdown, but a significant increase in memory usage. Further work on how to informatively select samples of $U$, \textit{e.g.} via importance sampling, could help alleviate this cost. 

\paragraph{Restrictive variational family.}
A second limitation is that, while our variational posterior is more expressive than the MF-GPLVM alternative, we still make strong assumptions on the form of our variational posterior. For example, we assume $q(x_n \mid U)$ is a normal distribution with a diagonal Gaussian covariance and $q(U) = \prod_{p=1}^P q(u_p)$. As a result, we expect that the A-S-GPLVM method proposed here still does not accurately match the exact posterior and further work on how to effectively relax those approximations may yield similar improvements in performance.

\section{Conclusion}
\label{sec:conclusion}

This work introduces SSVI for GPLVMs by coupling the variational posterior of the latent variables to the inducing variables. Across Oilflow, qPCR, MNIST, and FMNIST, the A-S-GPLVM achieves statistically significant improvements in train and test CRPS over the A-MF-GPLVM. Its RMSE and NLPD improvements are also statistically significant, except for RMSE on Oilflow.

Compared with IWVI-GPLVM, the A-S-GPLVM achieves significantly lower train and test NLPD on all four datasets, and significantly lower RMSE and CRPS on MNIST and FMNIST. Results on the smaller datasets are more mixed, and empirical 95\% coverage does not show uniform improvement. Together, these findings provide quantitative evidence that structured posterior dependence improves the quality of the learned predictive distribution. The qualitative uncertainty plots additionally show a more rapid increase in uncertainty away from the observed data (Figure~\ref{fig:latent-slice}), consistent with our motivation to better capture epistemic uncertainty in the learned data manifold.


Future work will explore how to use the improved uncertainty of the data manifold. For example, as a method of improving the sharing of information and uncertainty estimates in multiobjective Bayesian optimization, as in \citet{sedgwick2025TransferlearningBayesian}.

\newpage

\section{Acknowledgments}

The authors acknowledge financial support from Eli Lilly and Co. and the Engineering and Physical Sciences Research Council of the UK via Prosperity Partnership (grant
number EP/T005556/1). RM holds concurrent appointments as a Professor at Imperial College London and as an Amazon Scholar. This paper describes work performed at Imperial College London and is not associated with Amazon.

\bibliographystyle{plainnat}
\bibliography{ref}

\newpage

\appendix

\input{appendix_elbo_derivation}

\input{appendix_possible_extension}
\input{appendix_oilflow_plots}
\input{appendix_qpcr_plots}
\input{appendix_mnist_plots}
\input{appendix_fmnist_plots}

\end{document}

%% file: appendix_elbo_derivation.tex
\section{Graphical Models}
\label{sec:graphical-models}

\begin{figure}[t]
  \centering
\begin{subfigure}[t]{\dimexpr\linewidth/2-0.7em\relax}
    \centering
    \resizebox{0.45\linewidth}{!}{%
      \begin{tikzpicture}[baseline, every node/.style={transform shape}]
        \tikzset{
          >=Stealth,
          latent/.style   ={circle,draw,minimum size=18pt,inner sep=0pt},
          observed/.style ={latent,fill=black!10},
          const/.style    ={inner sep=0pt},
          plate/.style    ={rectangle,draw,rounded corners,inner sep=8pt}
        }
        
        \node[latent]   (x) at (0.5,0) {$x_n$};
        \node[observed] (y) at (2.2,0) {$y_n$};
        
        \node[latent]   (f) at ($(x)!0.5!(y)+(0,1.9)$) {$f_p$};
        \node[latent]   (u) at ($(f)+(0,1.9)$) {$u_p$};
        
        \node[const]    (theta) at (3.4,3.8) {$\theta$};
        \node[const]    (md)    at (-0.4,3.8) {$\mu_p$};
        \node[const]    (Sd)    at (-0.4,3.0) {$\Sigma_p$};
        \node[const]    (noise) at (3.2,0.15) {$\sigma_y^2$};
        \node[const]    (g)   at (0,-1.2) {$g$};
        \node[const]    (Z)     at ($(u)+(0,1.1)$) {$Z$};
        
        \draw[->] (x) -- (f);
        \draw[->] (u) -- (f);
        \draw[->] (f) -- (y);
        \draw[->] (theta) -- (u);
        \draw[->] (theta) -- (f);
        \draw[->] (md) -- (u);
        \draw[->] (Sd) -- (u);
        \draw[->] (noise) -- (y);
        \draw[->] (Z) -- (u);
        
        \draw[dashed,->] (g) -- (x);
        \draw[dashed,-] (y) to[bend left=35] (x);
        
        \node[plate, fit=(x)(y), label={[anchor=south east]south east:$N$}] (plateN) {};
        \node[plate, fit=(u)(f)(md)(Sd),
              label={[anchor=south east]south east:$P$}] (plateDF) {};
      \end{tikzpicture}
    }
  \end{subfigure}\hfill
  \begin{subfigure}[t]{\dimexpr\linewidth/2-0.7em\relax}
    \centering
    \resizebox{0.45\linewidth}{!}{%
      \begin{tikzpicture}[baseline, every node/.style={transform shape}]
     \tikzset{
          >=Stealth,
          latent/.style   ={circle,draw,minimum size=18pt,inner sep=0pt},
          observed/.style ={latent,fill=black!10},
          const/.style    ={inner sep=0pt},
          plate/.style    ={rectangle,draw,rounded corners,inner sep=8pt}
        }
        
        \node[latent]   (x) at (0.5,0) {$x_n$};
        \node[observed] (y) at (2.2,0) {$y_n$};
        
        \node[latent]   (f) at ($(x)!0.5!(y)+(0,1.9)$) {$f_p$};
        \node[latent]   (u) at ($(f)+(0,1.9)$) {$u_p$};
        
        \node[const]    (theta) at (3.4,3.8) {$\theta$};
        \node[const]    (md)    at (-0.4,3.8) {$\mu_p$};
        \node[const]    (Sd)    at (-0.4,3.0) {$\Sigma_p$};
        \node[const]    (noise) at (3.2,0.15) {$\sigma_y^2$};
        \node[const]    (h)   at (0,-1.2) {$h$};
        \node[const]    (Z)     at ($(u)+(0,1.1)$) {$Z$};
        
        \draw[->] (x) -- (f);
        \draw[->] (u) -- (f);
        \draw[->] (f) -- (y);
        \draw[->] (theta) -- (u);
        \draw[->] (theta) -- (f);
        \draw[->] (md) -- (u);
        \draw[->] (Sd) -- (u);
        \draw[->] (noise) -- (y);
        
        \draw[->] (Z) -- (u);
        
        \draw[dashed,->] (h) -- (x);
        \draw[dashed,-] (y) to[bend left=35] (x);
        \draw[dashed,-] (u) to[bend right=35] (x);

        \node[plate, fit=(x)(y), label={[anchor=south east]south east:$N$}] (plateN) {};
        \node[plate, fit=(u)(f)(md)(Sd),
              label={[anchor=south east]south east:$P$}] (plateDF) {};
      \end{tikzpicture}
    }
  \end{subfigure}

  \caption{Graphical models comparing the GPLVM variants.
Left: Amortized mean-field (A-MF-GPLVM), where the variational posteriors of the latent variables, $x_n$, are independent of the inducing point values, $u_p$.
Right: Amortized Structured (A-S-GPLVM, \textit{ours}), which conditions the value of the latent variable variational posterior on the global inducing variables, capturing dependencies between local and global variational factors. The dashed line indicates a variable used for amortization.}
  \label{fig:gplvm-three}
\end{figure}
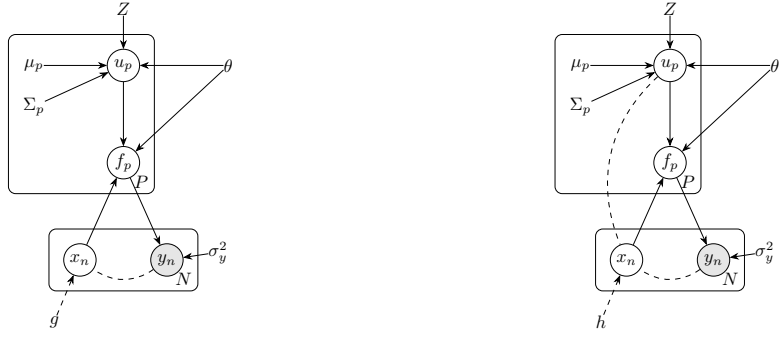

\section{SSVI ELBO derivation details}
\label{sec:elbo-derivation-details}

As stated in Section~\ref{sec:method}, we approximate the ELBO using $K$ Monte Carlo
samples $U_k \sim q(U)$, yielding
\begin{equation}
\label{eq:struc-elbo-appendix}
\mathcal L_S
\approx
\frac{1}{K}\sum_{k=1}^K \sum_{n=1}^{N} \hat{\mathcal L}_n(U_k)
- D_{\mathrm{KL}}\!\big(q(U)\,\|\,p(U)\big).
\end{equation}

To calculate the local ELBO, it is helpful to explicitly split the calculation across the $P$ output dimensions. Recalling the conditional relationships from Equation~\eqref{eq:conditional_f}, we can see that for one latent variable $x_n$ and inducing point sample for a single output $u_{pk} \in \R^{M}$, then the function output for a single observation and output, $f_{npk}\in\R$, the conditional distribution is given by
\begin{equation}
p(f_{npk} \mid u_{pk}, x_n) = \mathcal{N}(f_{npk} \mid \mu^{f}_{npk}, \sigma^{f}_{npk})
\end{equation}
where $\mu^{f}_{npk}, \sigma^f_{npk}\in \mathbb R$ are given by 
\begin{align}
\mu^{f}_{npk}
&= k_\theta(x_n, Z)^\top K_{ZZ}^{-1} u_{pk},
\label{eq:mu-f}
\\
\sigma^{f}_{npk}
&= k_\theta(x_n, x_n)
- k_\theta(x_n, Z) K_{ZZ}^{-1} k_\theta(x_n, Z)^\top .
\label{eq:sigma-f}
\end{align}
Note that $\sigma^{f}_{npk}$ is the variance of the distribution, but we have not included the squared term to keep the notation cleaner. Additionally, $p(y_n \mid f_n, x_n, U_k) = p(y_n|f_n)$ so we can further simplify our notation here by dropping $x_n$ and $U_k$ from the notation of the likelihood. 

The local ELBO can then be written as a sum of the $P$ expected log-likelihood terms  
\begin{equation}
    \fit_{npk} = \mathbb E_{p(f_{npk}\mid x_n,U_{k})\,q(x_n\mid U_k)}
\!\left[ \log p\!\left(y_{np} \mid f_{npk}\right)\right]
\end{equation}
and a KL divergence to give
\begin{equation}
\begin{aligned}
\hat{\mathcal L}_n(U_k)
&=
\sum_{p=1}^P \fit_{npk}
- D_{\mathrm{KL}}\!\left(q(x_n\mid U_k)\,\|\,p(x_n)\right).
\end{aligned}
\end{equation}

If the variational posterior is constrained to a parametric form, as it is in our case, the KL term $\KL(q(x_n\mid U)\|p(x_n))$ is available in closed form.

Proceeding in a very similar way to the MF-GPLVM, we can take the expectation of the log-likelihood over $p(f_{npk}| x_n, U_k)$, giving
\begin{equation}
\label{eq:expectation-wrt-f}
\fit_{npk}
=
- \frac{1}{2} \log(2\pi \sigma^2)
- \frac{1}{2 \sigma^2}
\, \mathbb{E}_{q(x_n \mid U_k)}
\Big[
(y_{np} - \mu^{f}_{npk})^2
+ \sigma^{f}_{npk}
\Big].
\end{equation}
We then define the so-called $\psi$-statistics, which are the required kernel expectations:
\begin{equation}
\begin{aligned}
\psi_{0n} &= \mathbb{E}_{q(x_n)}[k_{\theta}(x_n, x_n)] \in \R , \\
\psi_{1n} &= \mathbb{E}_{q(x_n)}[k_{\theta}(x_n, Z)] \in \R^{1 \times M}, \\
\psi_{2n} &= \mathbb{E}_{q(x_n)}[k_{\theta}(x_n, Z)^\top k_\theta(x_n, Z)] \in \R^{M \times M}.
\end{aligned}
\end{equation}
We can separate the two terms in the latent variable expectation in Equation~\eqref{eq:expectation-wrt-f}. The expectation of $\sigma^f_{npk}$ with respect to $q(x_n|U_k)$ is 
\begin{equation}
\hat\sigma^{f}_{npk} = \psi_{0n} - \operatorname{tr}\!\big(K_{ZZ}^{-1} \psi_{2n}\big).
\end{equation}
Expanding the quadratic expectation term gives
\begin{equation}
\mathbb{E}_{q(x_n| U_k)}\!\big[(y_{np} - \mu^{f}_{npk})^2\big]
= y_{np}^2
- 2y_{np}\,\mathbb{E}_{q(x_n| U_k)}[\mu^{f}_{npk}]
+ \mathbb{E}_{q(x_n| U_k)}\!\big[(\mu^{f}_{npk})^2\big].
\end{equation}
For the ARD kernel we use, the expectations have closed-form solutions \citep{titsias2010bayesian}.

Substituting the kernel expectations

\begin{equation}
\begin{aligned}
\mathbb{E}_{q(x_n \mid U_k)}\!\big[\mu^{f}_{npk}\big]
&= \psi_{1n} K_{ZZ}^{-1} u_{pk},
\\
\mathbb{E}_{q(x_n \mid U_k)}\!\big[(\mu^{f}_{npk})^2\big]
&= u_{pk}^\top K_{ZZ}^{-1} \psi_{2n} K_{ZZ}^{-1} u_{pk}.
\end{aligned}
\end{equation}

For notational convenience, we use
\begin{equation}
\hat\mu^{f}_{npk} = \mathbb{E}_{q(x_n|U_k)}[\mu^{f}_{npk}]
\qquad
\end{equation}

The expectations above were done for the $p$th output dimension of $y_n$. As a result, for all dimensions of $y_n$ we obtain: 
\begin{equation}
\sum_{p=1}^P \fit_{npk}
=  
-\frac{P}{2}\log(2\pi \sigma^2)
 -\frac{1}{2\sigma^2}\sum_{p=1}^P \Big[ \hat\sigma^{f}_{npk}
+ y_{np}^2
- 2y_{np}\,\hat\mu^{f}_{npk}
+ u_{pk}^\top K_{ZZ}^{-1}\psi_{2n}K_{ZZ}^{-1}u_{pk}
\Big].
\end{equation}

\section{Amortized Encoder}

As noted by \citet{hoffman2015ssvi}, the function $h(\cdot, \cdot)$ can be defined implicitly as $h(y_n, U) = \arg\max_{m_n, S_n} \hat \elbo_n(U)$, however, this requires solving a new optimization problem for $q(x_n|U)$ at each training step and for each sample of $U$. To avoid this expensive step, amortized inference can be used. In our testing, we found several design choices improved performance, which we detail below. 

\paragraph{Encoder architecture.}
In our experiments, we used two separate two-hidden-layer multilayer perceptrons (MLPs), one to encode the mean and one to encode the log-variance. The rationale for this was that the magnitudes of variation in optimal means and variances differ significantly.

\paragraph{Reparameterization of inducing variables.}
Both of these networks received the same inputs. The goal of these networks is to couple the information from the observation to the values of the inducing variables to improve the representation of the latent space. The naive way to do this would be to pass in the tuple $(U_k, y_n)$ for each observation and sample. While this approach is possible, we found that at late stages of training, when the variational distribution of the inducing variables, $q(U)$, had a high degree of certainty, all of the samples of the inducing variables $U_k$ lie very close to one another. In order to give a better-conditioned input to the network, while preserving the information required to couple the inducing points to the observations, we use the reparameterization trick to sample $U_k$. 

For each sample of $u_{pk} \in \R^{M}$, we sample a standard normal variable 
\(
\xi_{pk} \sim \mathcal{N}(0, I)
\)
which we transform to be a sample from the variational posterior, $q(u_p) = \mathcal{N}(\mu_p, \Sigma_p)$ using
\begin{equation}
u_{pk} = \mu_p + L_p \xi_{pk},
\end{equation}
where $L_p$ is the Cholesky factor of $\Sigma_{p}$.
The main benefit of the reparameterization trick is that it allows gradients to propagate through samples of the global inducing variables $U$. However, as an added benefit, the reparameterized input $\Xi_k$ is well spread out over the input space, while maintaining all of the sample-specific information required to couple the inducing sample to the output and condition the latent variable on the inducing point samples.
Therefore, we pass the variable $\Xi_k = \{\xi_{pk}\}_{p=1}^P$.

\paragraph{Local variational posterior.}
We additionally apply a soft-plus transform to ensure that the variance of the variational posterior is strictly positive, so the local variational posterior becomes 
\begin{align}
\mu_{nk} &= h_\mu(y_n, \Xi_k), \\
\log \sigma_{nk} &= h_\sigma(y_{n}, \Xi_k), \\
\Sigma_{nk} &= \text{diag}(\mathrm{softplus}(\log \sigma_{nk}) + \epsilon_{\mathrm{min}}),
\end{align}
where $\epsilon_{\mathrm{min}}$ is a small positive constant ensuring numerical stability.
These parameters define the local variational distribution
\(
q(x_n \mid U_k) = \mathcal{N}(\mu_{nk}, \Sigma_{nk}).
\)

\paragraph{Annealed prior regularization.}
When training on larger datasets, we found that the mean encoder network was slow to converge to making predictions with variance approximately equal to one. To speed up this convergence, we adapted our local variable by scaling the KL regularization term up by a factor of $\beta$ larger than one at the start of training, similar to the $\beta$-VAE \citep{higgins2017beta}, so the local ELBO looks like
\begin{equation}
\label{eq:annealed-local-variable}
\hat{\mathcal L}_n^{\beta}(U_k)
= \mathbb{E}_{p(f_n \mid x_n, U_k)\, q(x_n \mid U_k)}\big[\log p(y_n \mid f_n, x_n, U_k)\big]
- \beta\, D_{\mathrm{KL}}\!\big(q(x_n \mid U_k)\,\|\,p(x_n)\big).
\end{equation}
Over the course of training, we decay this variable from an initial value $\beta>1$ to $\beta=1$ at the end of training. We applied this only to the MNIST/FMNIST datasets, with an initial value of $\beta = 10$ and a quadratic decay rate to $\beta=1$. It is important to emphasize that this variable does not change the objective at the end of training; its only function is to speed up convergence to the optimal variational posterior. 

\paragraph{Comparison to mean field amortization.}
In amortized mean field GPLVMs, the encoder takes only $y_n$ as input and produces a variational posterior $q(x_n)$ independent of the inducing variables.

\section{Model Configuration}
\label{sec:model-config}

For the smaller datasets (oil flow, qPCR), the hyperparameters are:
\begin{itemize}
    \item Latent Dimensionality ($Q$): 10 dimensions
    \item Inducing Points ($M$): 32 per output dimension
    \item Training Iterations: 100{,}000 for all four models
    \item Batch Size: 32
    \item Random Seeds: 0,1,2,3,4
    \item Initial lengthscales of $1.0$ and observation noise variance of $1.0$
    \item Number of samples from the inducing variables ($K$): 64
    \item Learning rate: $1e^{-3}$
    \item MLP hidden layer dims: $(128, 64)$
\end{itemize}

For the larger datasets (MNIST, FMNIST), the hyperparameters are:
\begin{itemize}
    \item Latent Dimensionality ($Q$): 10 dimensions
    \item Inducing Points ($M$): 64 per output dimension
    \item Training Iterations: 200{,}000 for all three models
    \item Batch Size: 32
    \item Random Seeds: 0,1,2,3,4
    \item Initial lengthscales of $1.0$ and observation noise variance of $1.0$
    \item Number of samples from the inducing variables ($K$): 4
    \item Learning rate: $1e^{-4}$
    \item MLP hidden layer dims: $(512, 256)$
\end{itemize}

\section{Computational setup}
\label{sec:computational-setup}

All experiments reported for Oilflow, qPCR, MNIST, and FMNIST using the hyperparameter configurations described above were run on similar GPU nodes in a high-performance computing cluster. The precise hardware specifications varied slightly across runs. Detailed measurements of the runtime and memory overhead of the proposed structured approximation, measured on a representative node, are provided in Appendix~\ref{sec:runtime-memory}.

%% file: appendix_possible_extension.tex
\begin{table}[H]
\centering
\small
\caption{Two-sided Welch's $t$-test $p$-values over $5$ independent runs comparing A-MF-GPLVM and A-S-GPLVM.}
\label{tab:significance_tests}

\begin{tabular}{lccccc}
\toprule
Dataset & RMSE (train) & RMSE (test) & NLPD (train) & NLPD (test) & ELBO \\
\midrule
Oilflow & $0.12$ & $0.55$ & $<10^{-3}$ & $<10^{-3}$ & $<10^{-4}$ \\
qPCR    & $<10^{-3}$ & $<10^{-3}$ & $<10^{-3}$ & $<10^{-3}$ & $<10^{-3}$ \\
MNIST   & $<10^{-4}$ & $<10^{-4}$ & $<10^{-4}$ & $<10^{-4}$ & $<10^{-4}$ \\
FMNIST  & $<10^{-4}$ & $<10^{-4}$ & $<10^{-4}$ & $<10^{-4}$ & $<10^{-4}$ \\
\bottomrule
\end{tabular}%
\end{table}

\section{Statistical significance of the observed improvements}
\label{sec:statistical-significance}

To assess whether the gains from the structured variational family are robust across random initialization and stochastic optimization noise, we performed a two-sided Welch's $t$-test over $5$ independent runs, comparing A-MF-GPLVM and A-S-GPLVM on each dataset. The resulting $p$-values are reported in Table~\ref{tab:significance_tests}.

Overall, the structured method yields statistically significant improvements on the probabilistic metrics across all datasets. In particular, for qPCR all reported differences are significant at the $10^{-3}$ level or smaller, and for MNIST and FMNIST all reported differences are significant at the $10^{-4}$ level or smaller. On Oilflow, both NLPD improvements remain highly significant ($p<10^{-3}$), and the ELBO difference is also significant ($p<10^{-4}$), whereas the RMSE differences are not statistically significant ($p=0.12$ on the training split and $p=0.55$ on the test split).

This distinction is important for interpreting the effect of the proposed method. In particular, the Oilflow result shows that the structured posterior can significantly improve the probabilistic quality of the model, as measured by NLPD and ELBO, even when the point-estimate accuracy measured by RMSE does not change significantly. This supports our claim that the main benefit of the structured variational family is in improving the posterior approximation and predictive uncertainty, rather than merely shifting the predictive mean.

\begin{table}[H]
\centering
\small
\caption{Empirical runtime and peak GPU memory over $100$ training iterations on Oilflow, using $B=32$, $M=32$, $Q=10$, $K=32$ and float64 precision.}
\label{tab:runtime_memory}
\begin{tabular}{lcc}
\hline
Metric & A-MF-GPLVM & A-S-GPLVM \\
\hline
Wall-clock time (s) & $5.03$ & $8.36$ \\
Peak GPU memory (MB) & $29.0$ & $598.7$ \\
\hline
\end{tabular}
\end{table}

\section{Runtime and memory overhead}
\label{sec:runtime-memory}

The proposed structured approximation introduces two main sources of overhead relative to A-MF-GPLVM. First, the encoder input becomes larger, since the sampled inducing-variable state is concatenated with the observation rather than processing the observation alone. Second, each ELBO evaluation uses $K$ Monte Carlo replicas. Both operations are parallelizable, so the primary practical trade-off is expected to be increased memory use rather than an inherently sequential increase in computation.

To quantify this overhead empirically, we measured runtime and peak GPU memory on the Oilflow dataset for $100$ training iterations using a minibatch size of $B=32$, $M=32$ inducing points, latent dimension $Q=10$, $12$ observed dimensions, $K=32$ Monte Carlo samples, and float64 precision throughout. The experiments were run on an NVIDIA RTX 2080Ti GPU (22\,GB VRAM) with an AMD EPYC 7282 CPU and CUDA 13.0. The results are shown in Table~\ref{tab:runtime_memory}.

In this configuration, A-S-GPLVM increases the wall-clock time from $5.03$\,s to $8.36$\,s over $100$ iterations, corresponding to an approximate $1.66\times$ increase. In terms of memory, peak GPU usage rises from $29.0$\,MB to $598.7$\,MB. This is consistent with the fact that the $K$ inducing-variable samples can be processed in parallel, so the dominant cost of the structured estimator is storing the replicated computation graph and intermediate activations.

We emphasize that these measurements characterize the present implementation rather than an unavoidable algorithmic constant. In our current codebase, the additional wall-clock overhead is partly influenced by implementation details in a shared layered design across models. The more stable conclusion is that the proposed method preserves the same sparse, minibatched stochastic variational structure as amortized mean-field GPLVMs, but pays a memory cost in exchange for a richer posterior approximation.

\section{Ablation studies}

\begin{figure}[H]
\centering

\begin{subfigure}{0.45\textwidth}
    \centering
    \includegraphics[width=\linewidth]{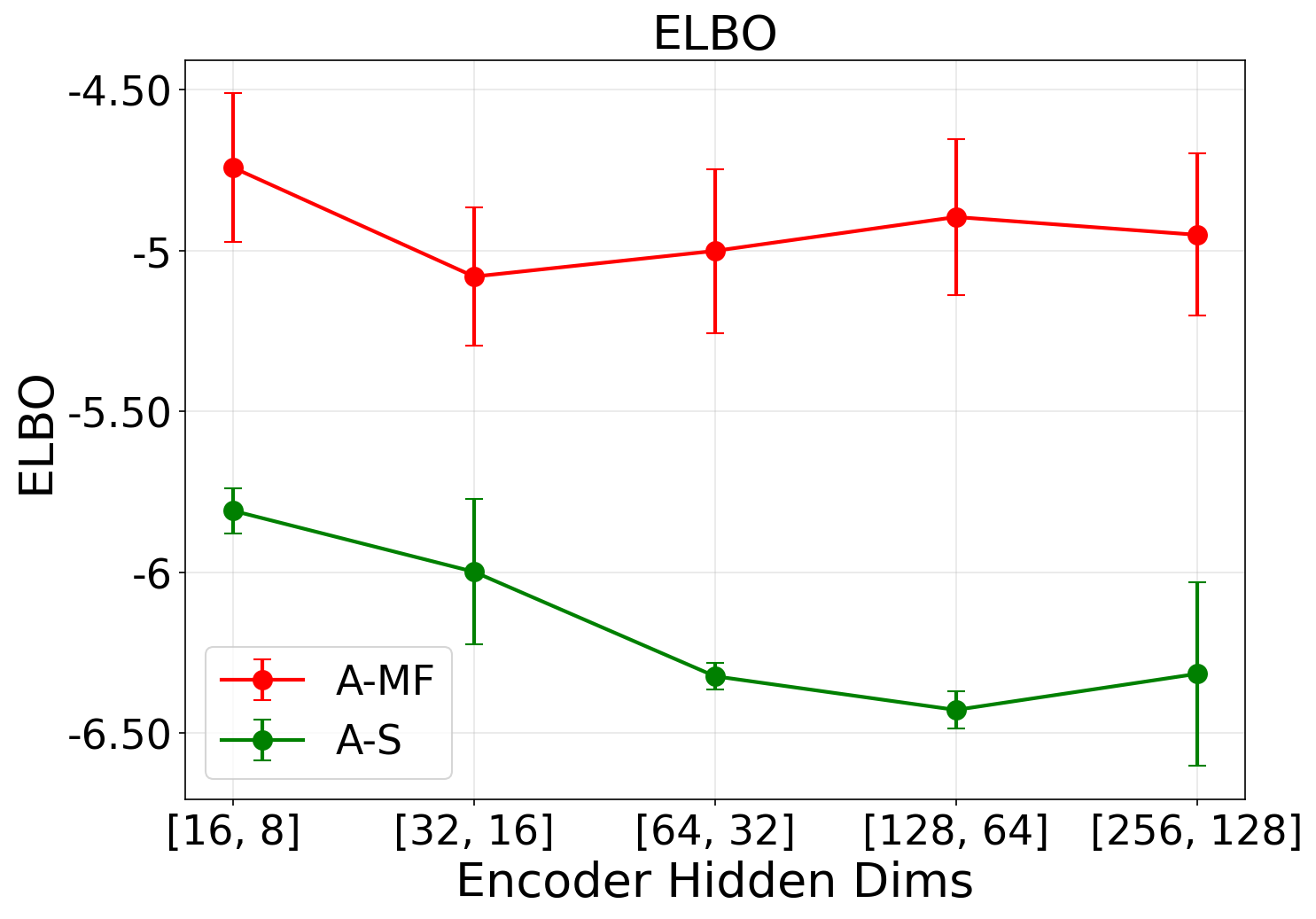}
    \caption{ELBO}
    \label{fig:ablation_encoder_elbo}
\end{subfigure}
\hfill
\begin{subfigure}{0.45\textwidth}
    \centering
    \includegraphics[width=\linewidth]{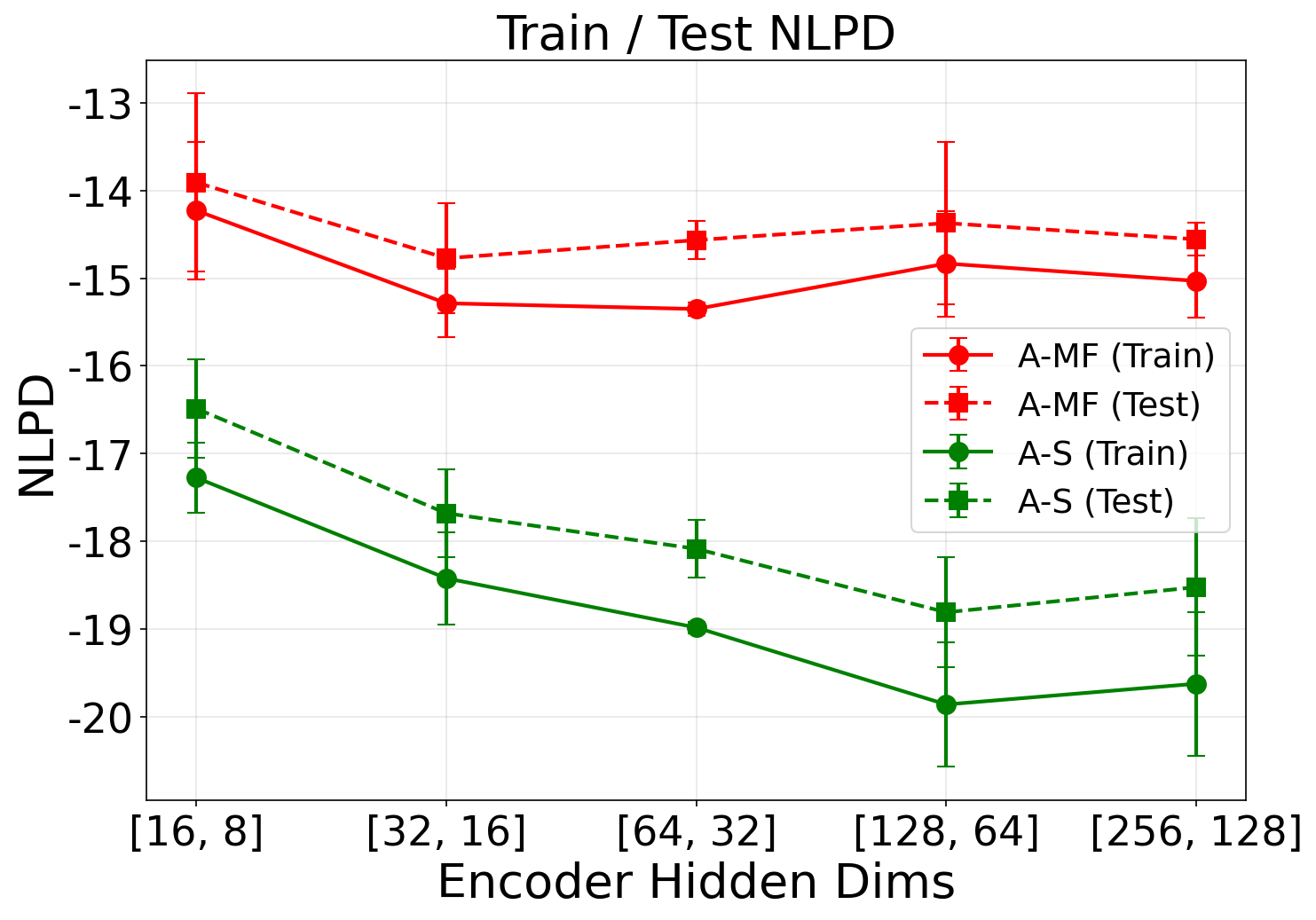}
    \caption{NLPD}
    \label{fig:ablation_encoder_nlpd}
\end{subfigure}

\vspace{0.8em}

\begin{subfigure}{0.45\textwidth}
    \centering
    \includegraphics[width=\linewidth]{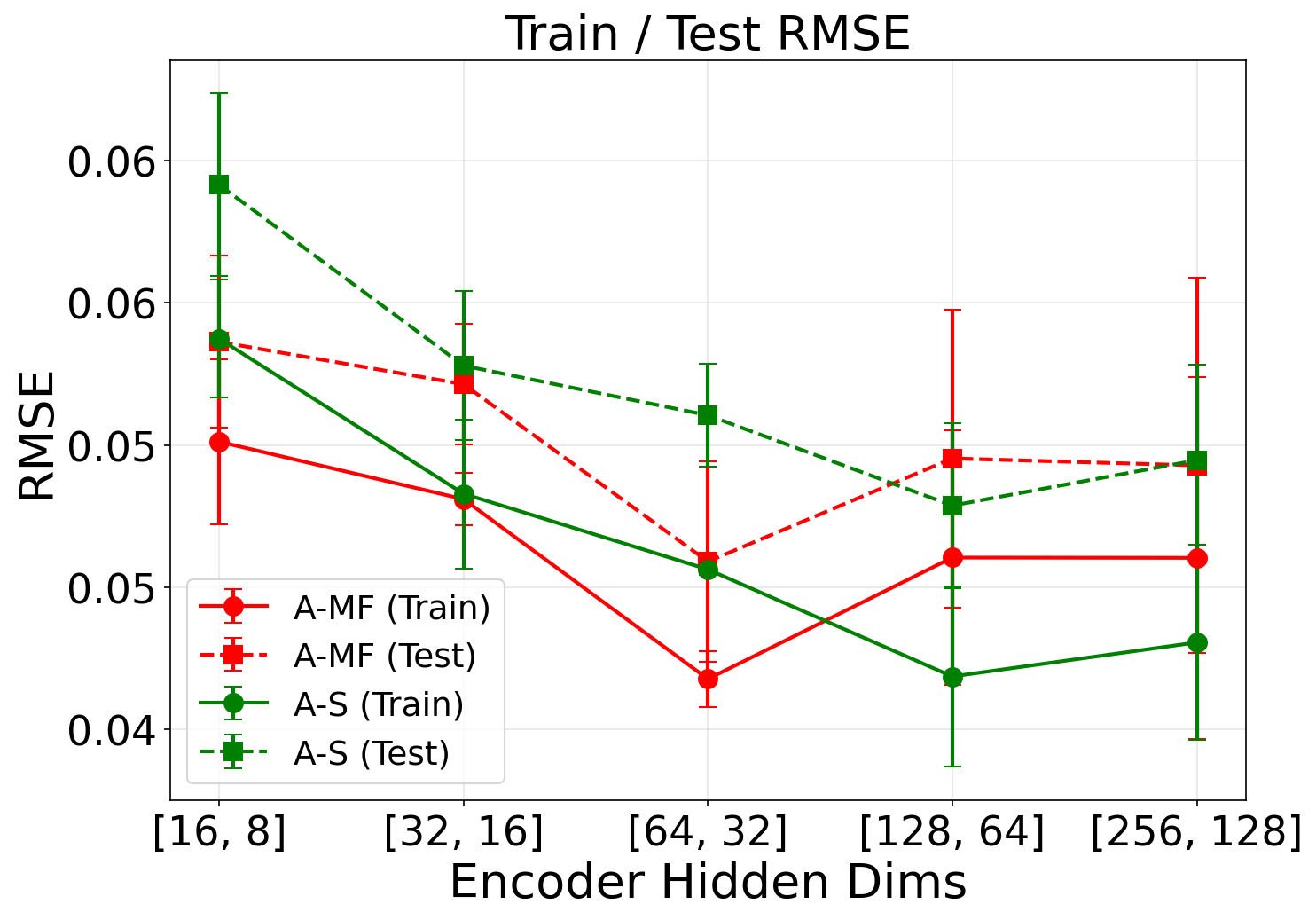}
    \caption{RMSE}
    \label{fig:ablation_encoder_rmse}
\end{subfigure}

\caption{Ablation over encoder architecture on the Oilflow dataset. We compare A-MF-GPLVM and A-S-GPLVM across different two-hidden-layer MLP encoder widths.}
\label{fig:ablation_encoder}
\end{figure}

\begin{figure}[H]
\centering

\begin{subfigure}{0.45\textwidth}
    \centering
    \includegraphics[width=\linewidth]{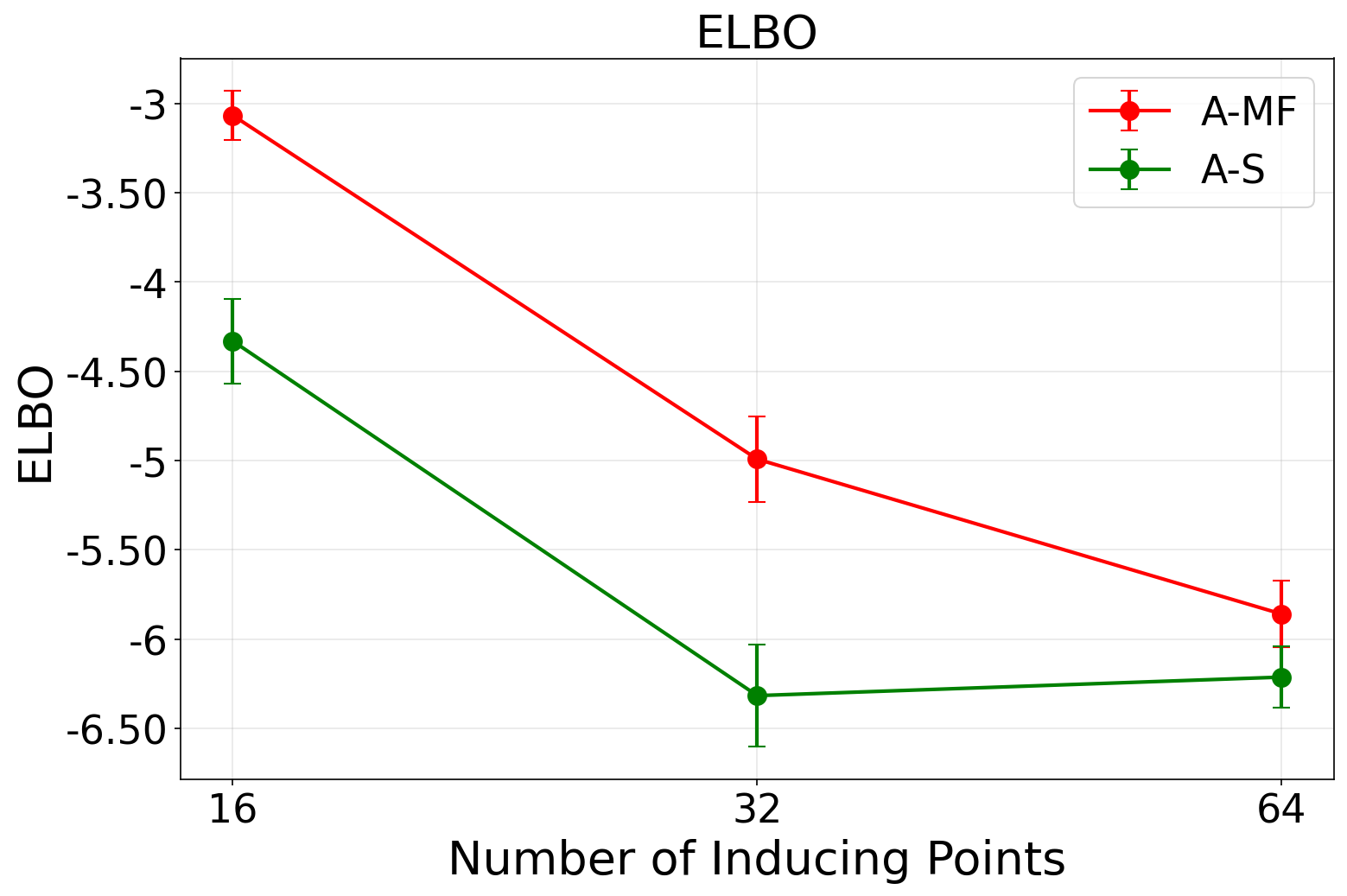}
    \caption{ELBO}
    \label{fig:ablation_inducing_elbo}
\end{subfigure}
\hfill
\begin{subfigure}{0.45\textwidth}
    \centering
    \includegraphics[width=\linewidth]{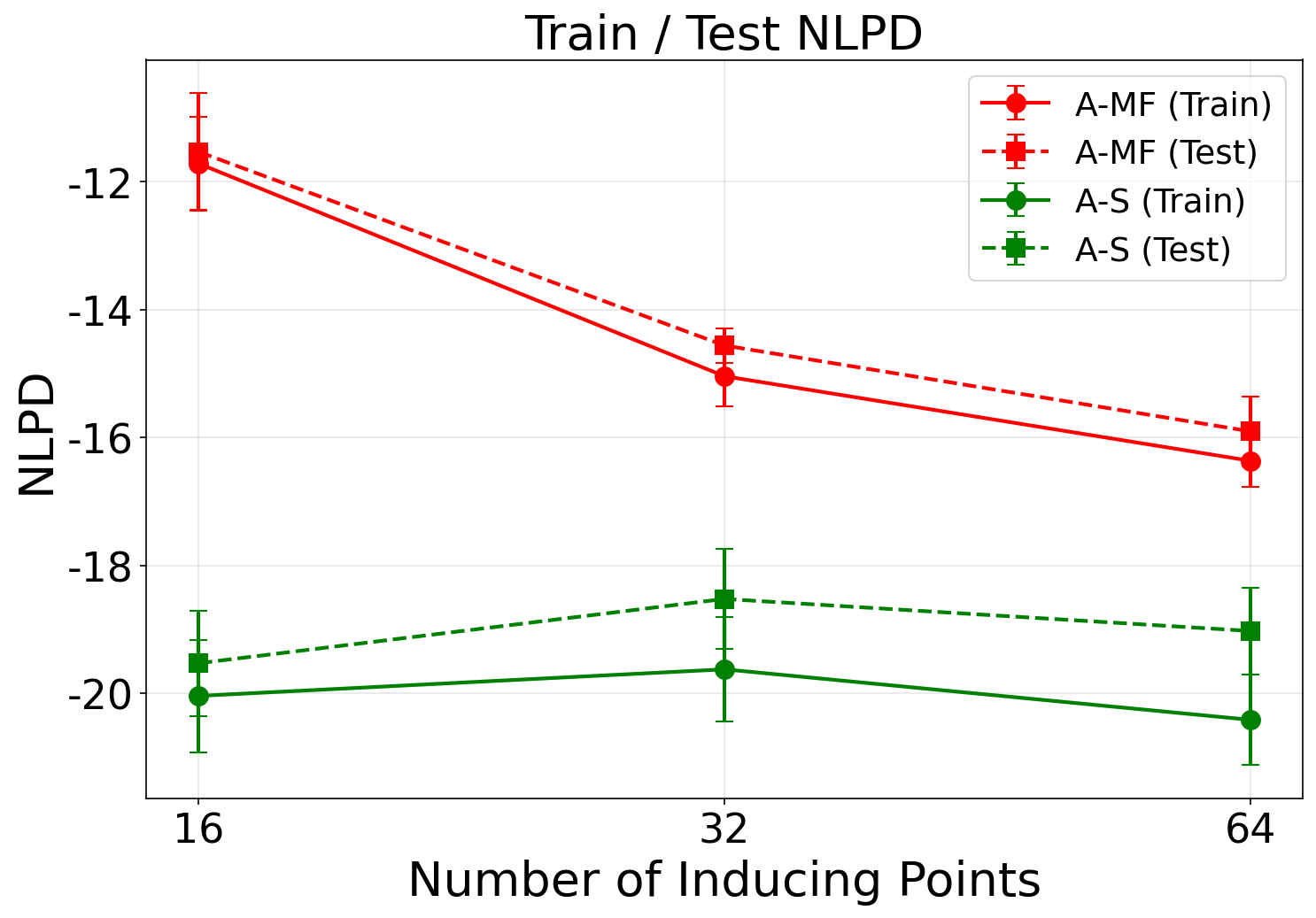}
    \caption{NLPD}
    \label{fig:ablation_inducing_nlpd}
\end{subfigure}

\vspace{0.8em}

\begin{subfigure}{0.45\textwidth}
    \centering
    \includegraphics[width=\linewidth]{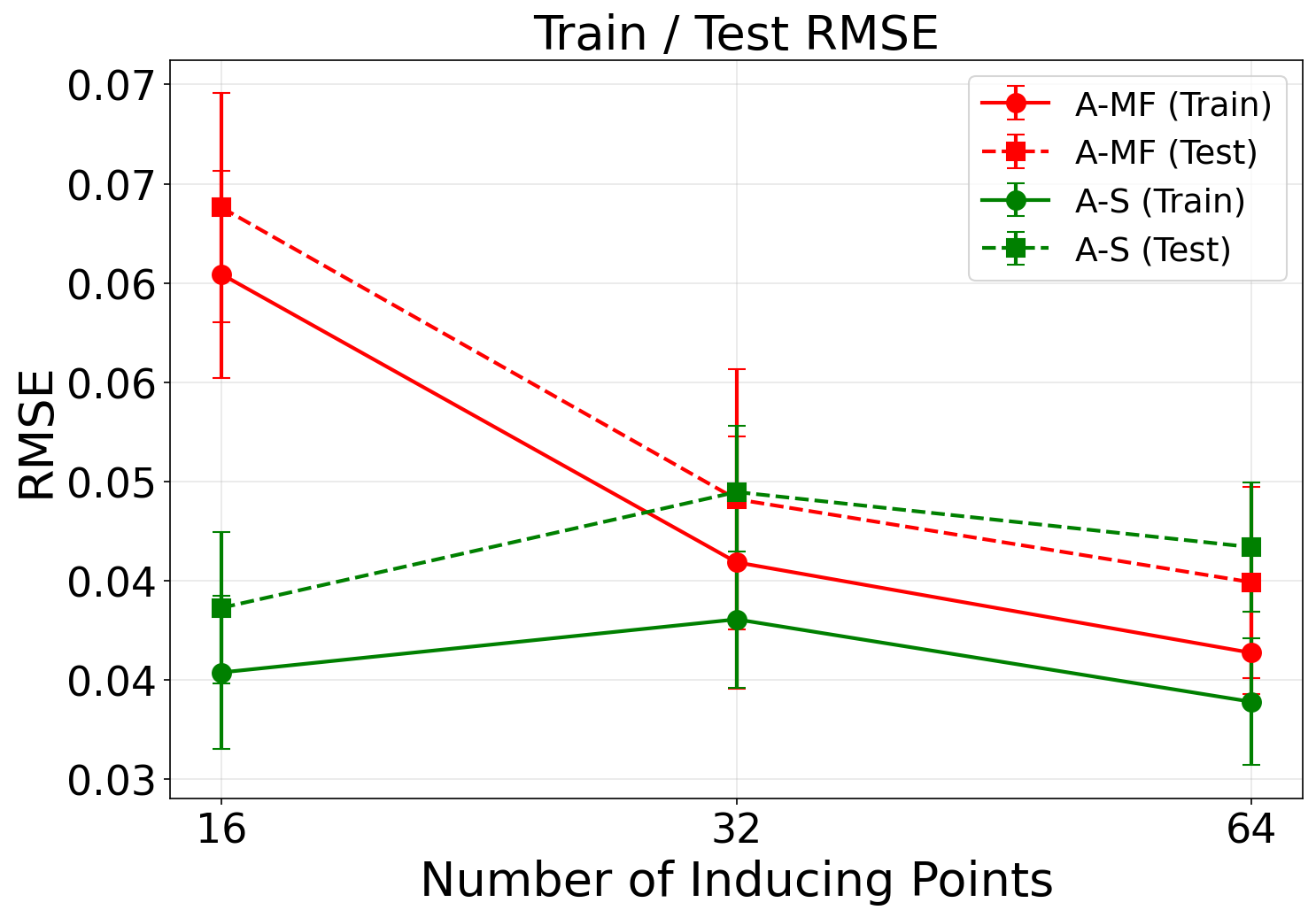}
    \caption{RMSE}
    \label{fig:ablation_inducing_rmse}
\end{subfigure}

\caption{Ablation over the number of inducing points on the Oilflow dataset. We compare A-MF-GPLVM and A-S-GPLVM for different values of the number of inducing points.}
\label{fig:ablation_inducing}
\end{figure}

\begin{figure}[H]
\centering

\begin{subfigure}{0.45\textwidth}
    \centering
    \includegraphics[width=\linewidth]{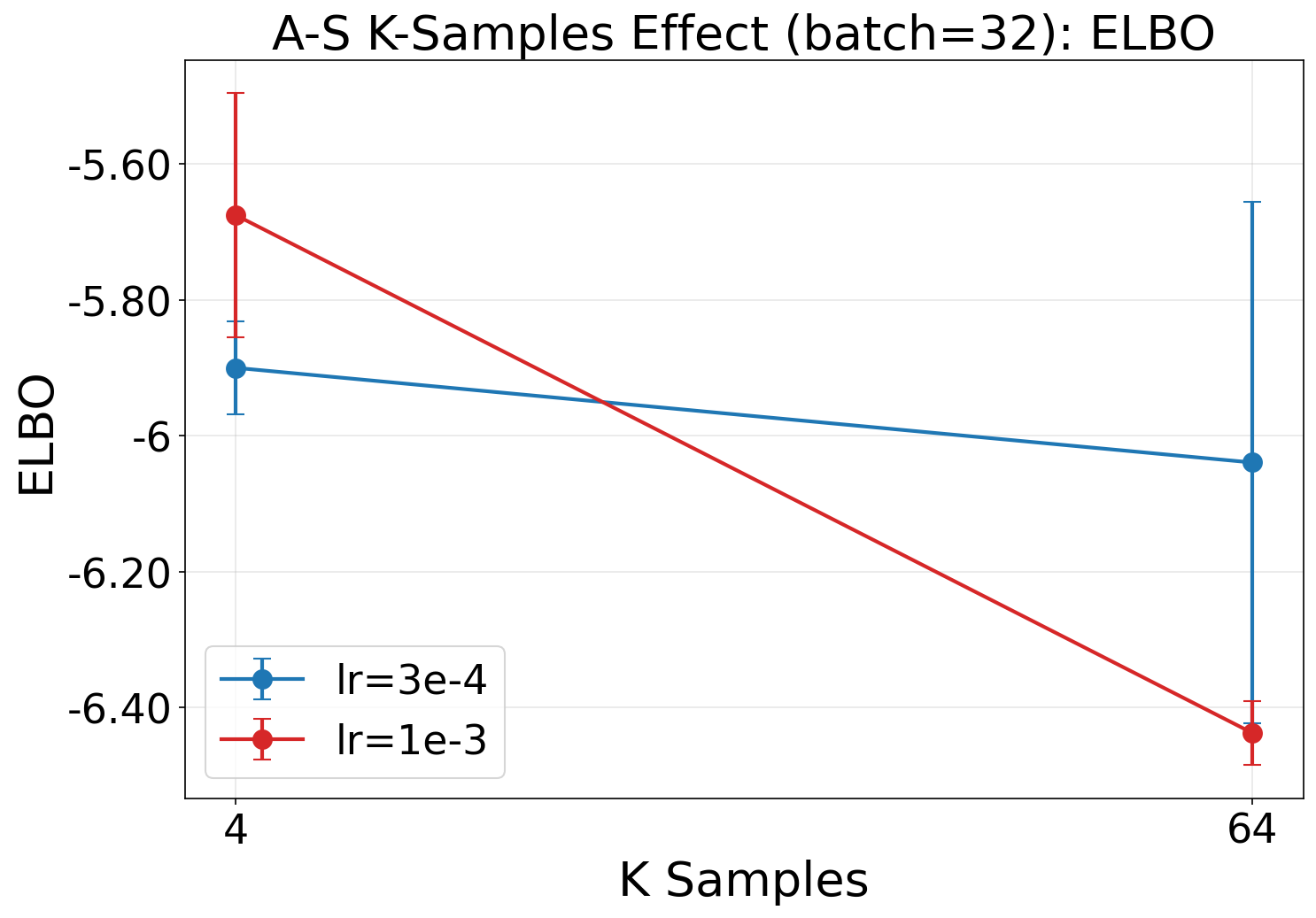}
    \caption{ELBO}
    \label{fig:ablation_k_elbo}
\end{subfigure}
\hfill
\begin{subfigure}{0.45\textwidth}
    \centering
    \includegraphics[width=\linewidth]{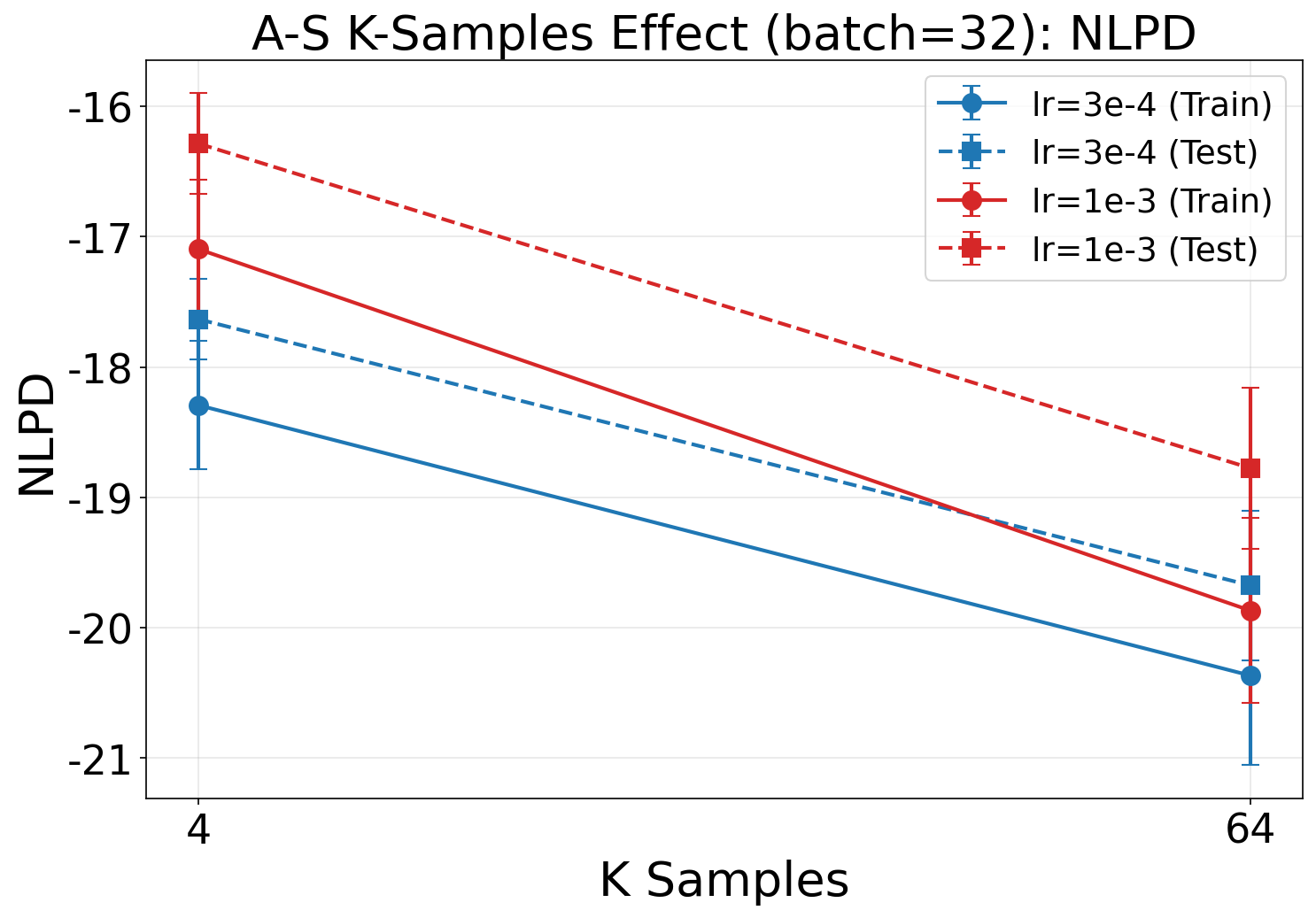}
    \caption{NLPD}
    \label{fig:ablation_k_nlpd}
\end{subfigure}

\vspace{0.8em}

\begin{subfigure}{0.45\textwidth}
    \centering
    \includegraphics[width=\linewidth]{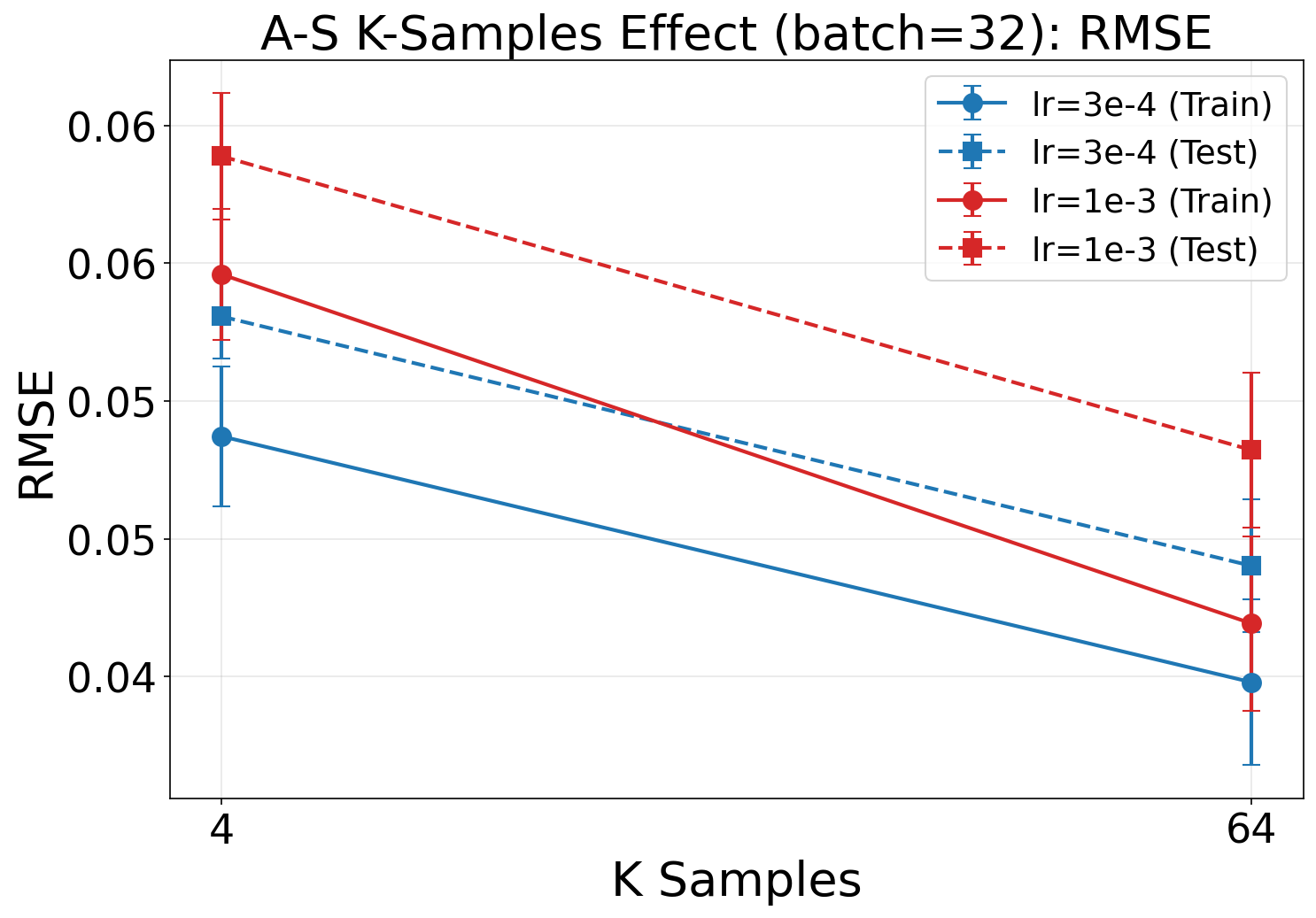}
    \caption{RMSE}
    \label{fig:ablation_k_rmse}
\end{subfigure}

\caption{Ablation over the number of sampled inducing-variable states on the Oilflow dataset.}
\label{fig:ablation_k}
\end{figure}

To better isolate the source of the improvement in A-S-GPLVM, we carried out ablations on the encoder architecture, the number of inducing points, and the number of sampled inducing-variable states. The corresponding plots are shown in Figures~\ref{fig:ablation_encoder}, \ref{fig:ablation_inducing}, and \ref{fig:ablation_k}.

\paragraph{Encoder architecture.}
We first varied the widths of the two-hidden-layer MLP encoder across the settings $[16,8]$, $[32,16]$, $[64,32]$, $[128,64]$, and $[256,128]$ for both A-MF-GPLVM and A-S-GPLVM. Across this range, the total encoder size did not have a strong effect on performance, while A-S-GPLVM consistently outperformed A-MF-GPLVM on the probabilistic metrics (ELBO and NLPD). This supports our claim that the main benefit does not come from increased encoder capacity. Increasing encoder size in the mean-field model only makes the deterministic map $y_n \mapsto q(x_n)$ more flexible, whereas the structured model can adapt the local posterior to the sampled global GP state through $q(x_n \mid U)$. In particular, even a small encoder with SSVI remains better than a larger encoder with MFVI.

\paragraph{Number of inducing points.}
We next fit the Oilflow dataset using $16$, $32$, and $64$ inducing points. Across all tested values, A-S-GPLVM performed better than A-MF-GPLVM on ELBO and NLPD, and often performed comparably to or better than A-MF-GPLVM with twice as many inducing points. We also observe that the margin between the two models becomes smaller with larger numbers of inducing points. In our current implementation, we suspect this happens because the encoder input dimension in A-S-GPLVM grows with the number of inducing points and becomes too large for the small MLP used in this ablation.

\paragraph{Number of inducing-variable samples.}
Finally, we varied the number of sampled inducing-variable states $K$. In the ablation plots, we compare $K=4$ and $K=64$, and also show the effect of using learning rates $10^{-3}$ and $3\times 10^{-4}$. Larger values of $K$ lead to better results, which is consistent with the reduced variance of the gradient estimator when more samples are used. In memory-constrained settings, however, increasing $K$ may not be possible. The ablation also shows that for small $K$, lowering the learning rate can recover performance by countering the increased gradient variance, at the cost of slower convergence.

Taken together, these ablations support the interpretation that the main gains come from the structured variational family rather than from increased encoder capacity, while also clarifying the dependence of performance on the number of inducing points and inducing-variable samples.

%% file: appendix_oilflow_plots.tex
\section{Additional Results Plots}
\label{sec:additional-plots}

We have included additional plots relating to the experiments described in Section~\ref{sec:experiments} here. 


\begin{figure*}[t]
  \centering
  \subcaptionbox{MF-GPLVM\label{fig:latent-slice-gauss-ap}}{%
    \includegraphics[width=0.9\linewidth]{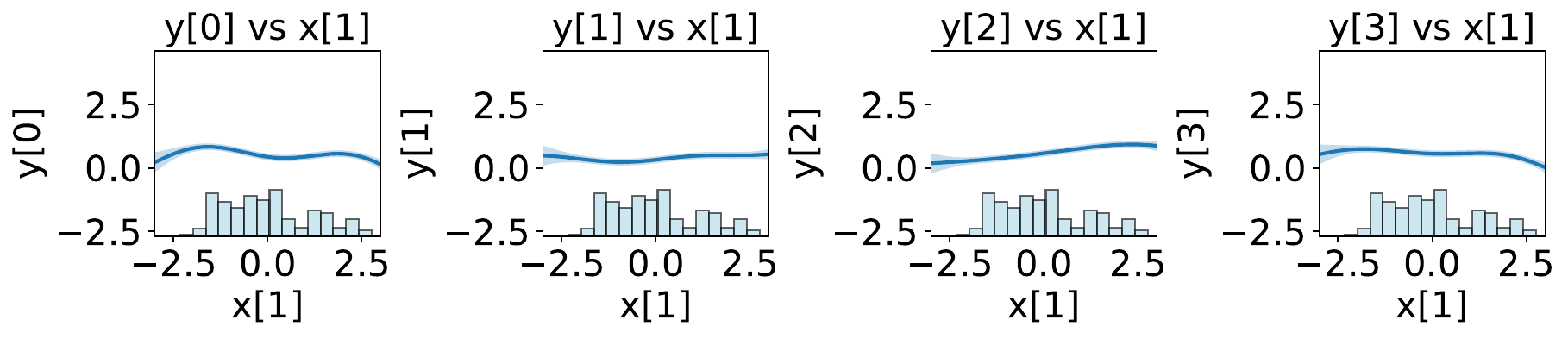}
  }\\[0.5em]
  \subcaptionbox{A-MF-GPLVM,\label{fig:latent-slice-nn-ap}}{%
    \includegraphics[width=0.9\linewidth]{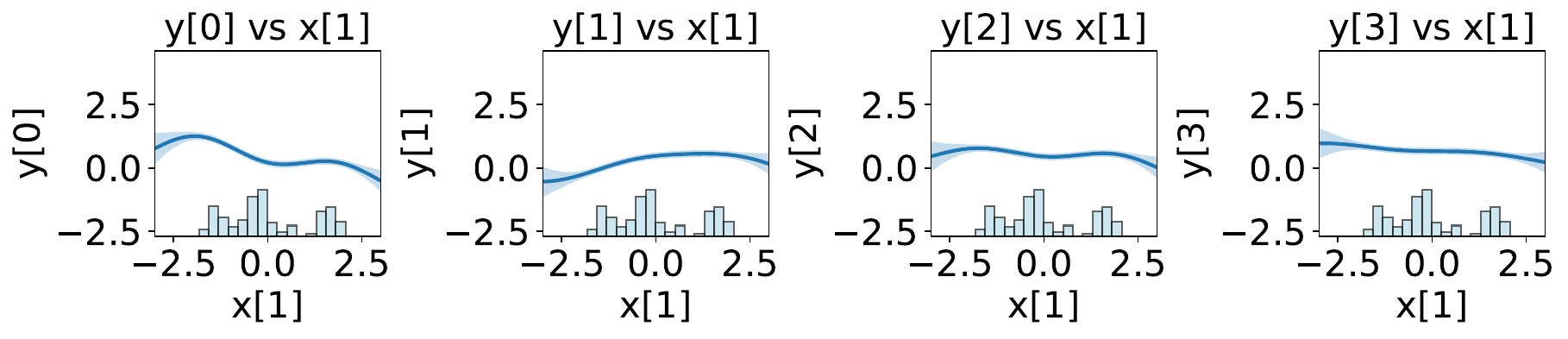}
  }\\[0.5em]
  \subcaptionbox{A-S-GPLVM\label{fig:latent-slice-ssvi-ap}}{%
    \includegraphics[width=0.9\linewidth]{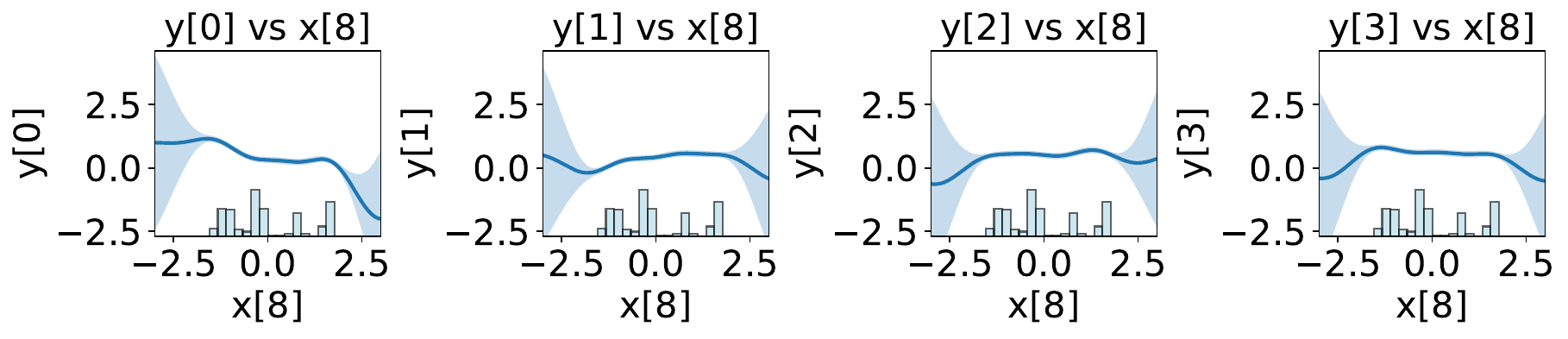}
  }

  \caption{Latent slice through the most informative latent dimension. 
  Curves show predictive means; shaded areas denote predictive uncertainty for each output. The A-S-GPLVM GP mapping has significantly larger predictive uncertainty away from the data.}
  \label{fig:latent-slice-oilflow-app}
\end{figure*}

%% file: appendix_qpcr_plots.tex
\label{sec:qpcr-restuls-appendix}

\begin{figure}[t]
  \centering
  \subcaptionbox{MF-GPLVM\label{fig:qpcr-results-gauss}}{%
    \includegraphics[width=0.45\textwidth, keepaspectratio]{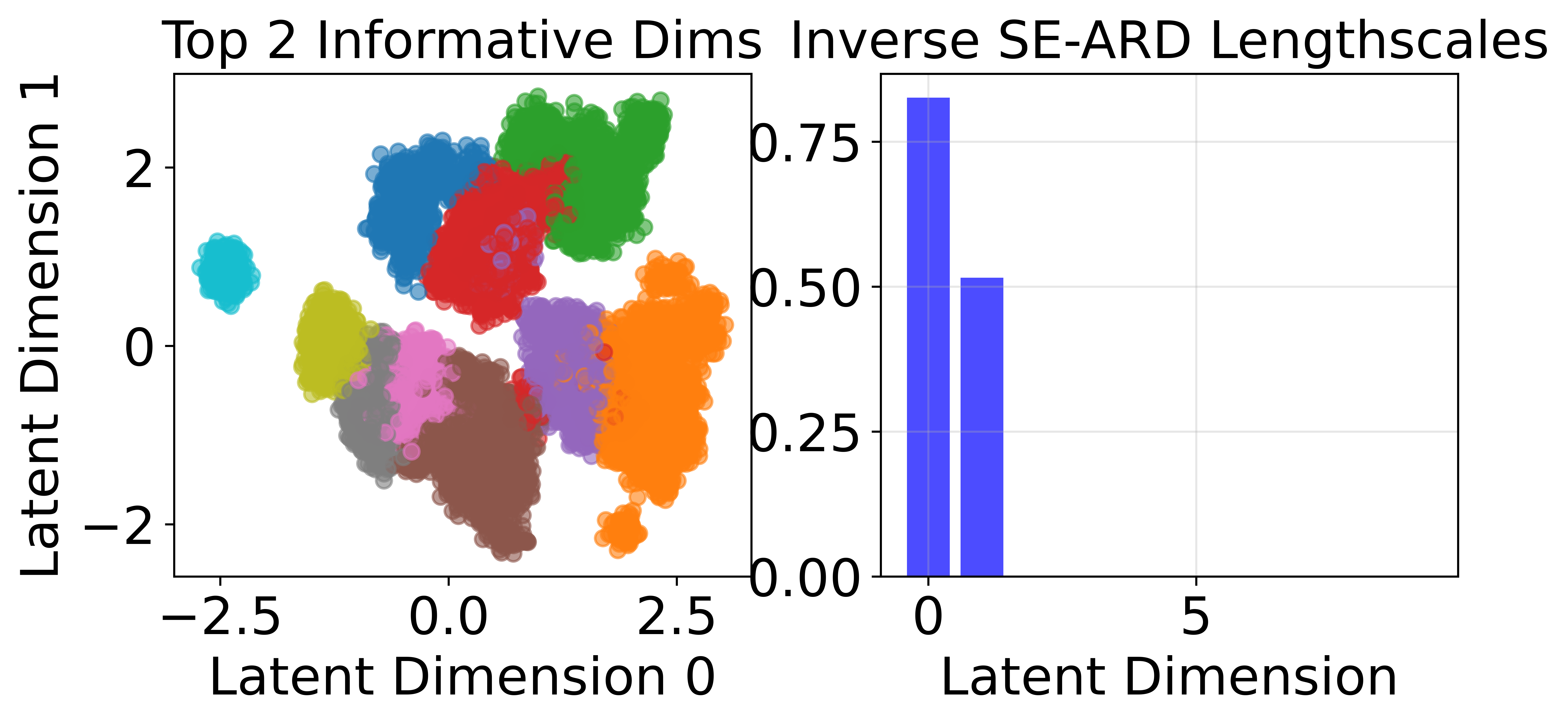}
  }\hfill
  \subcaptionbox{A-MF-GPLVM\label{fig:qpcr-results-nn}}{%
    \includegraphics[width=0.45\textwidth, keepaspectratio]{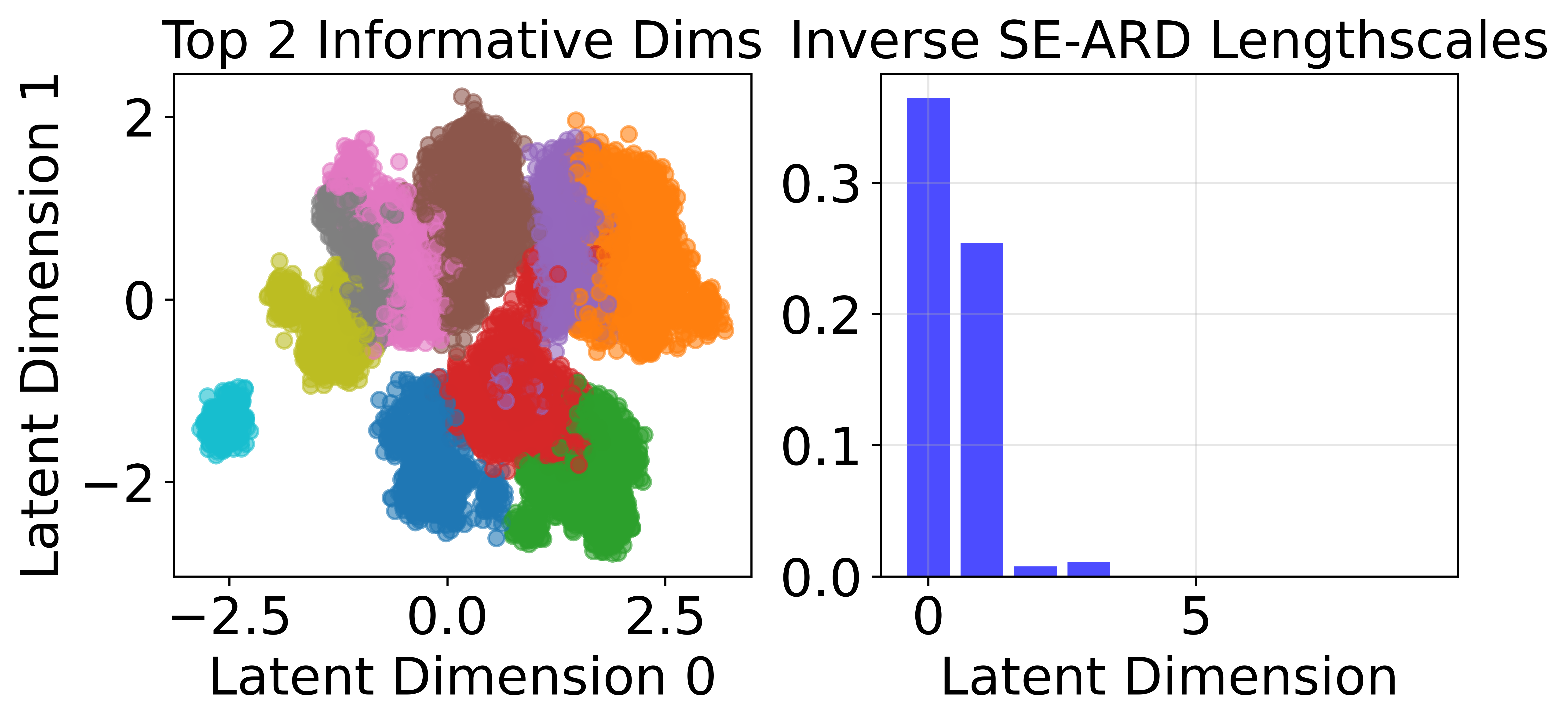}
  }\hfill
  \subcaptionbox{A-S-GPLVM\label{fig:qpcr-results-ssvi}}{%
    \includegraphics[width=0.45\textwidth, keepaspectratio]{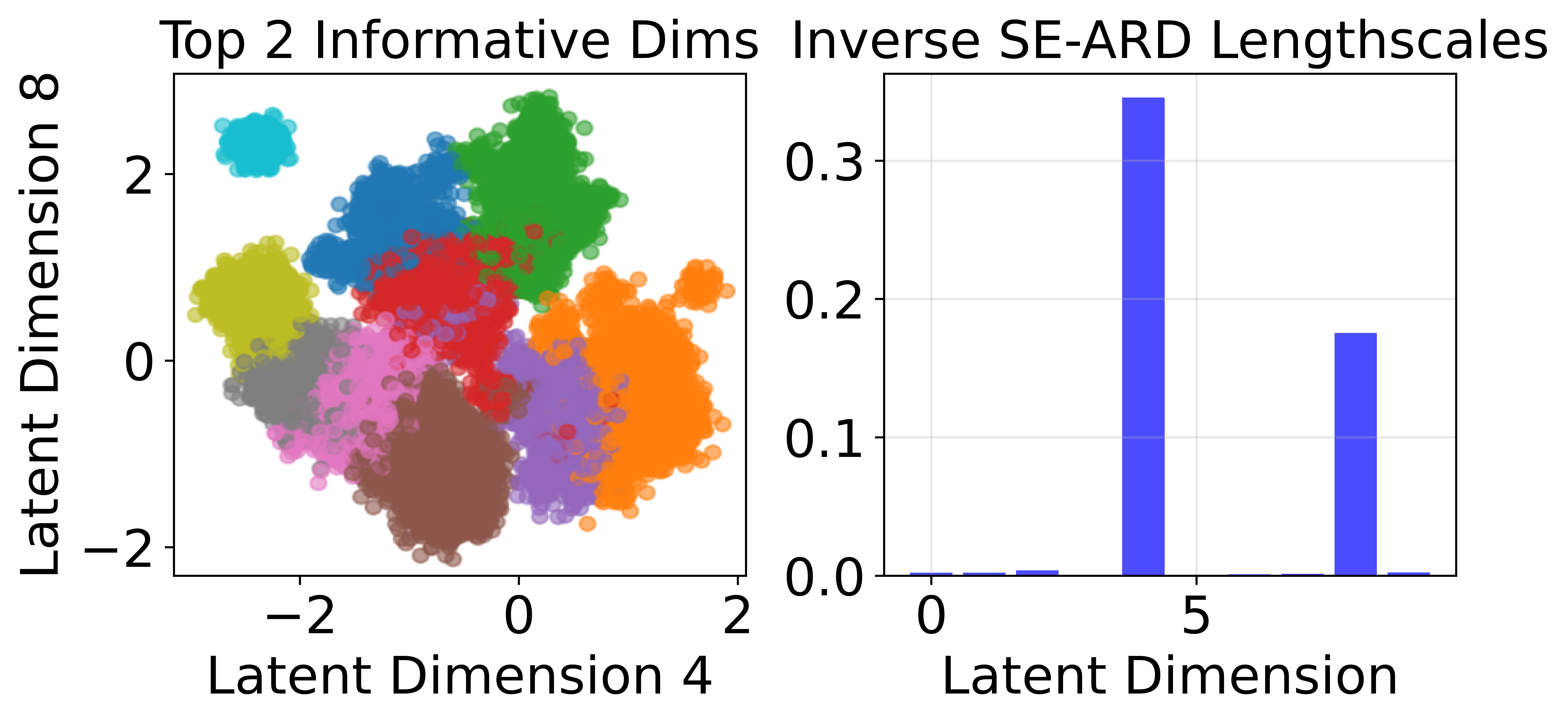}
  }

  \caption{
  Latent space comparison on the qPCR dataset, for (a) MF-GPLVM, (b) A-MF-GPLVM, (c) A-S-GPLVM.  
  }
  \label{fig:qpcr-results-app}
\end{figure}


\begin{figure*}[t]
    \centering
    \includegraphics[width=\linewidth]{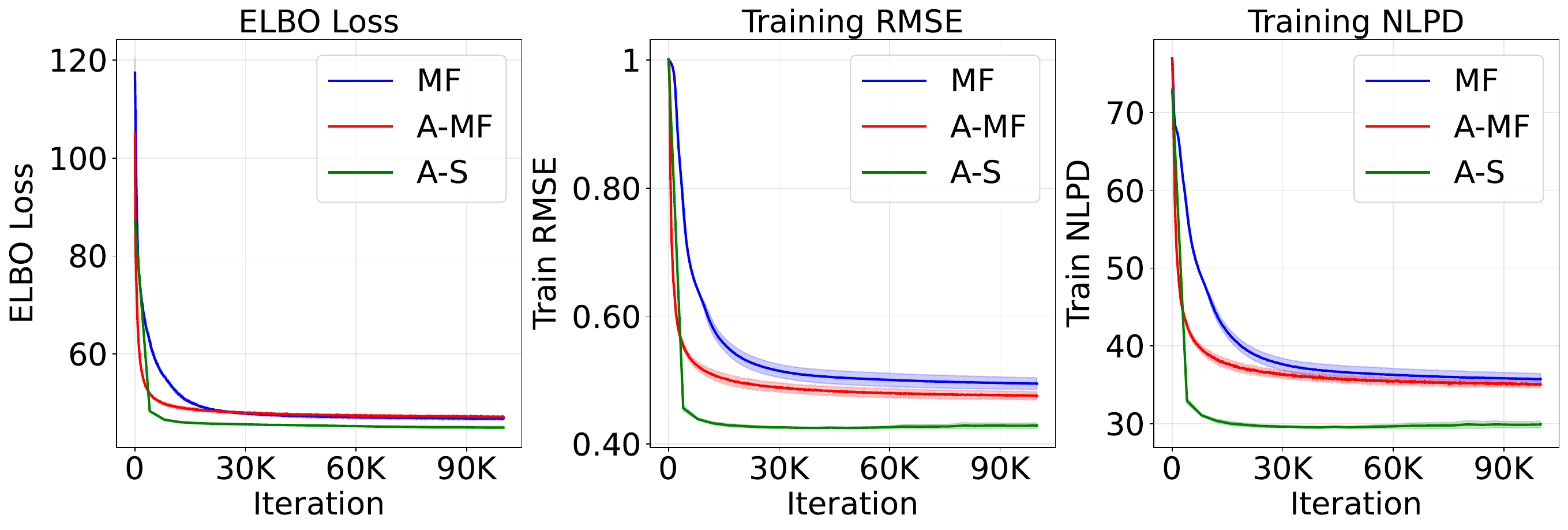}
    \caption{Training dynamics comparison across inference methods on the qPCR dataset. 
    The panels show the ELBO, RMSE, and NLPD trajectories over training iterations. 
    A-S-GPLVM converges to a lower final NLPD than MF-GPLVM and A-MF-GPLVM.}
    \label{fig:qpcr-metrics}
\end{figure*}

%% file: appendix_mnist_plots.tex
\label{sec:mnist-restuls-appendix}

\begin{figure}[t]
\centering
\begin{subfigure}[t]{\linewidth}
    \centering
    \includegraphics[width=\linewidth]{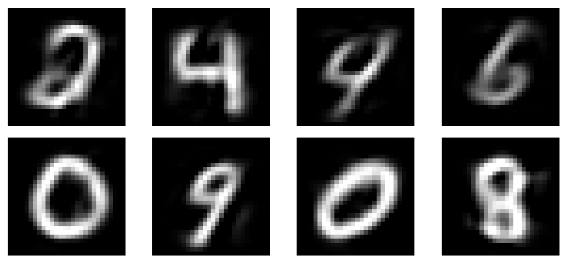}
    \subcaption{MNIST}
    \label{fig:mnist-mf-reconstruction}
\end{subfigure}

\begin{subfigure}[t]{\linewidth}
    \centering
    \includegraphics[width=\linewidth]{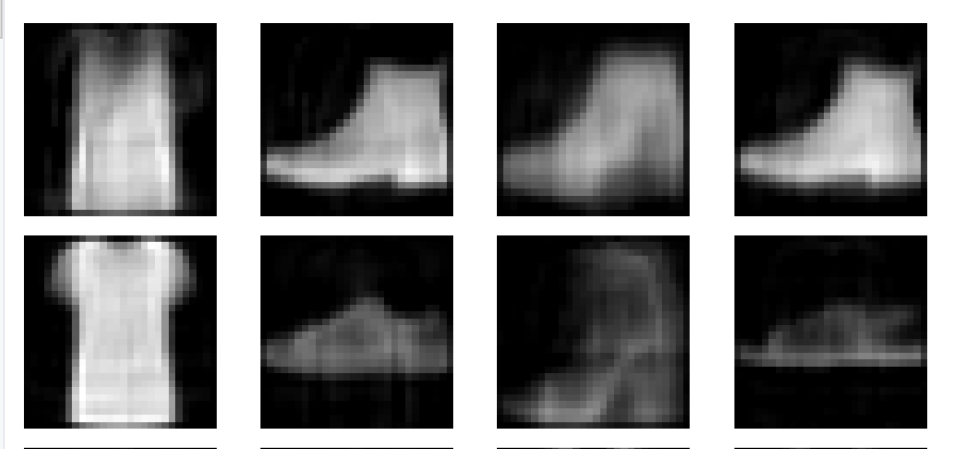}
    \subcaption{FMNIST}
    \label{fig:fmnist-mf-reconstruction}
\end{subfigure}
\caption{Sample reconstructions using the A-MF-GPLVM. Each image shows the posterior predictive mean of the reconstructed observation.}
\label{fig:mf-reconstructions}
\end{figure}

\begin{figure}[t]
    \centering
    \begin{subfigure}[b]{\linewidth}
        \centering
        \includegraphics[width=\linewidth]{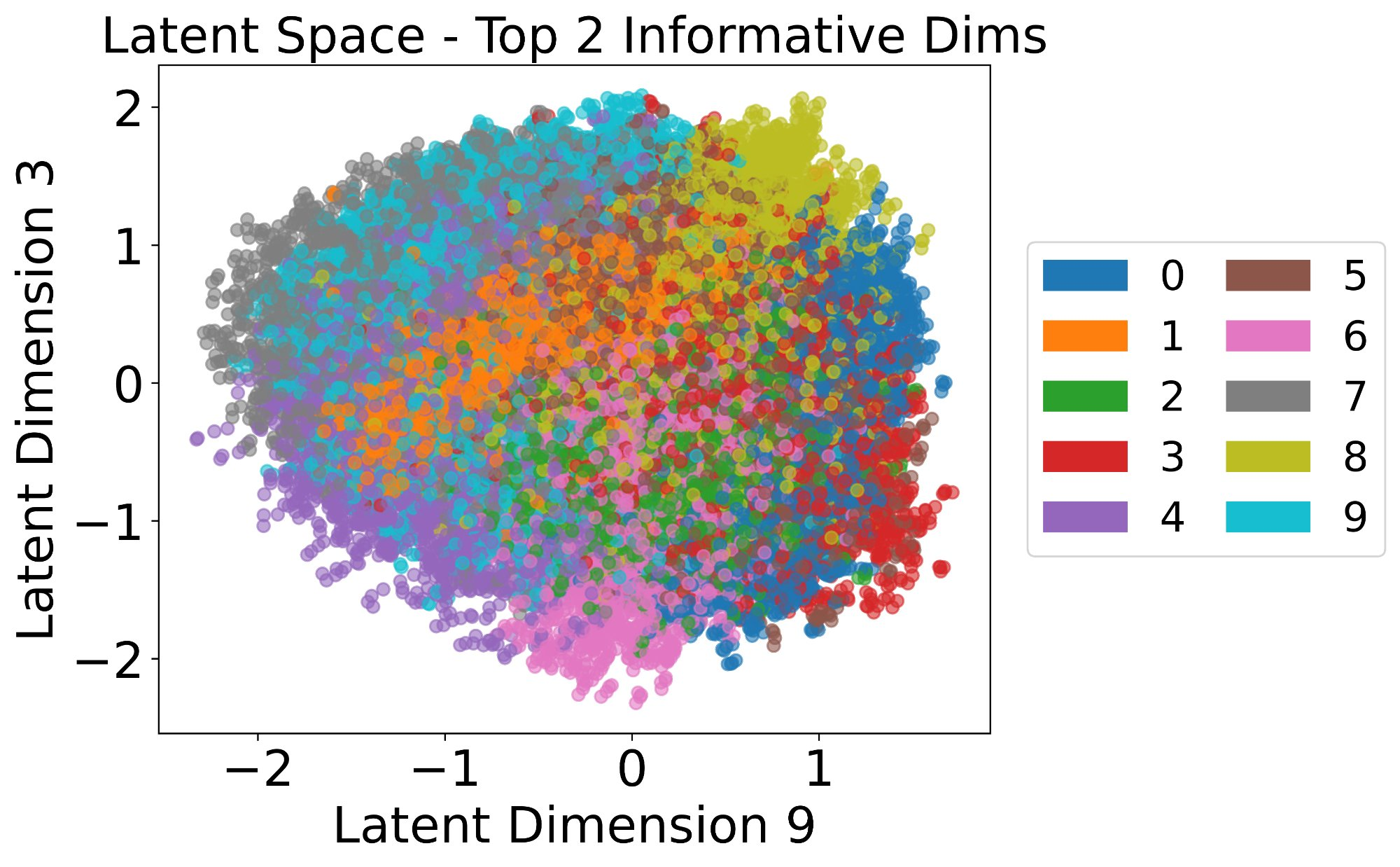}
        \caption{A-S-GPLVM}
        \label{fig:mnist-ssvi-latent}
    \end{subfigure}
    \hfill
    \begin{subfigure}[b]{\linewidth}
        \centering
        \includegraphics[width=\linewidth]{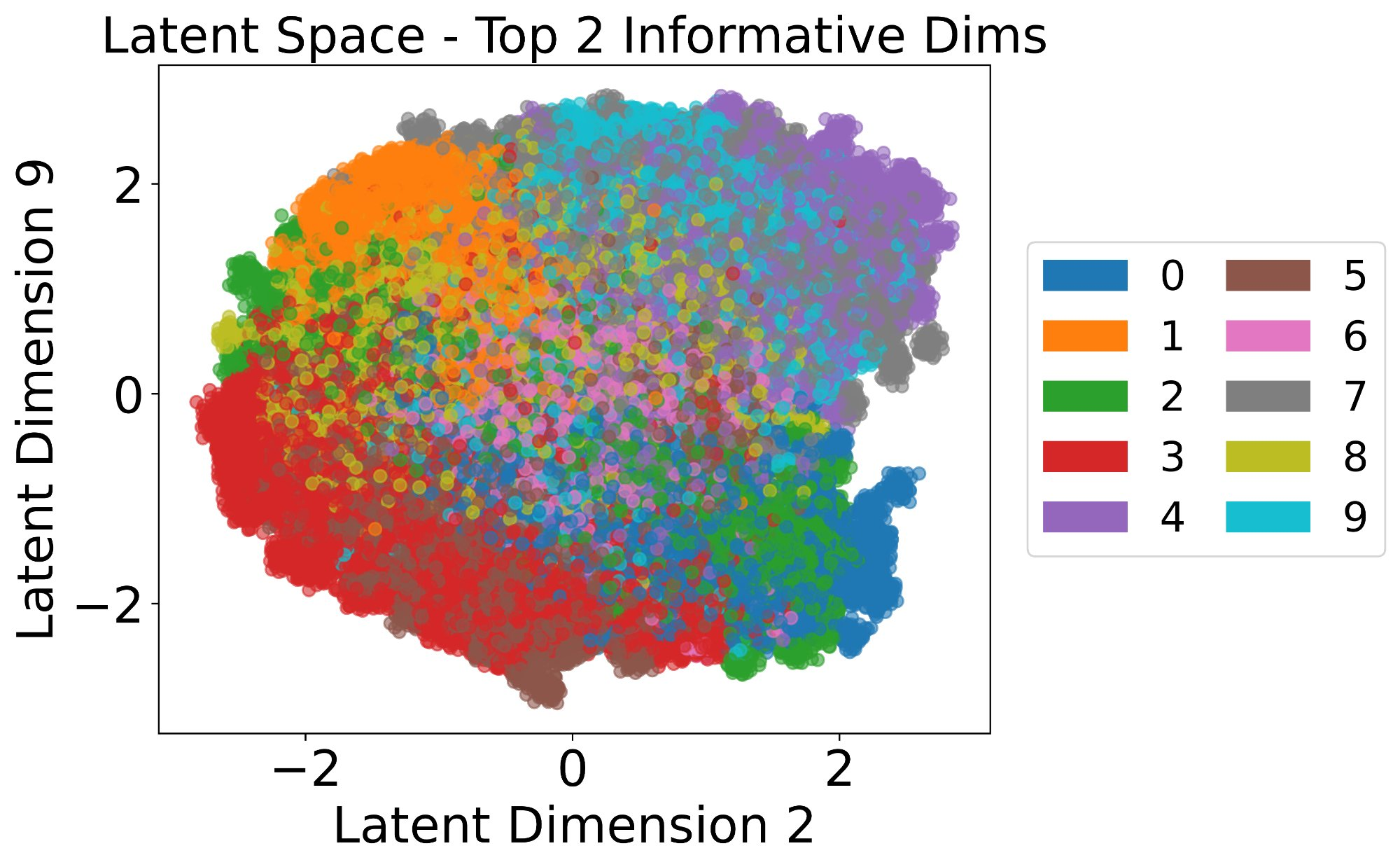}
        \caption{A-MF-GPLVM}
        \label{fig:mnist-amf-latent}
    \end{subfigure}
    \caption{Latent spaces for the MNIST dataset.}
    \label{fig:mnist-latent-space}
\end{figure}

\begin{figure*}[t]
    \centering
    \includegraphics[width=\linewidth]{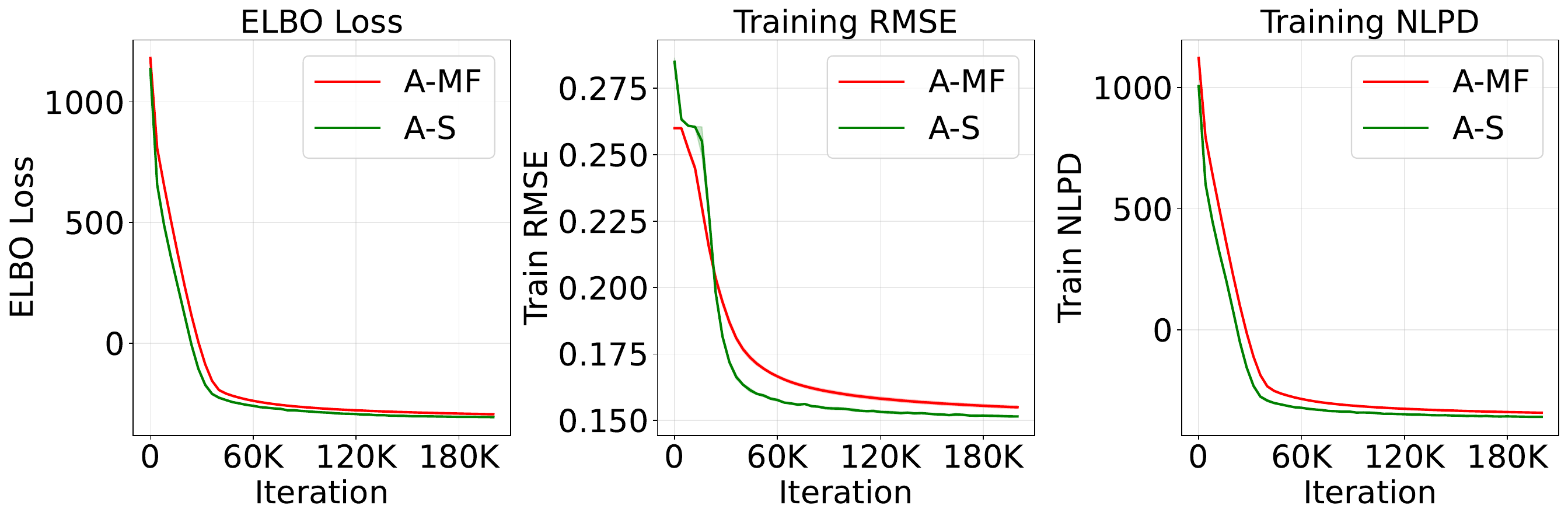}
    \caption{Training dynamics comparison across inference methods on the MNIST dataset. 
    The panels show the ELBO, RMSE, and NLPD trajectories over training iterations. 
    }
    \label{fig:mnist-metrics}
\end{figure*}

%% file: appendix_fmnist_plots.tex
\label{sec:fmnist-restuls-appendix}

\begin{figure*}[t]
    \centering
    \includegraphics[width=\linewidth]{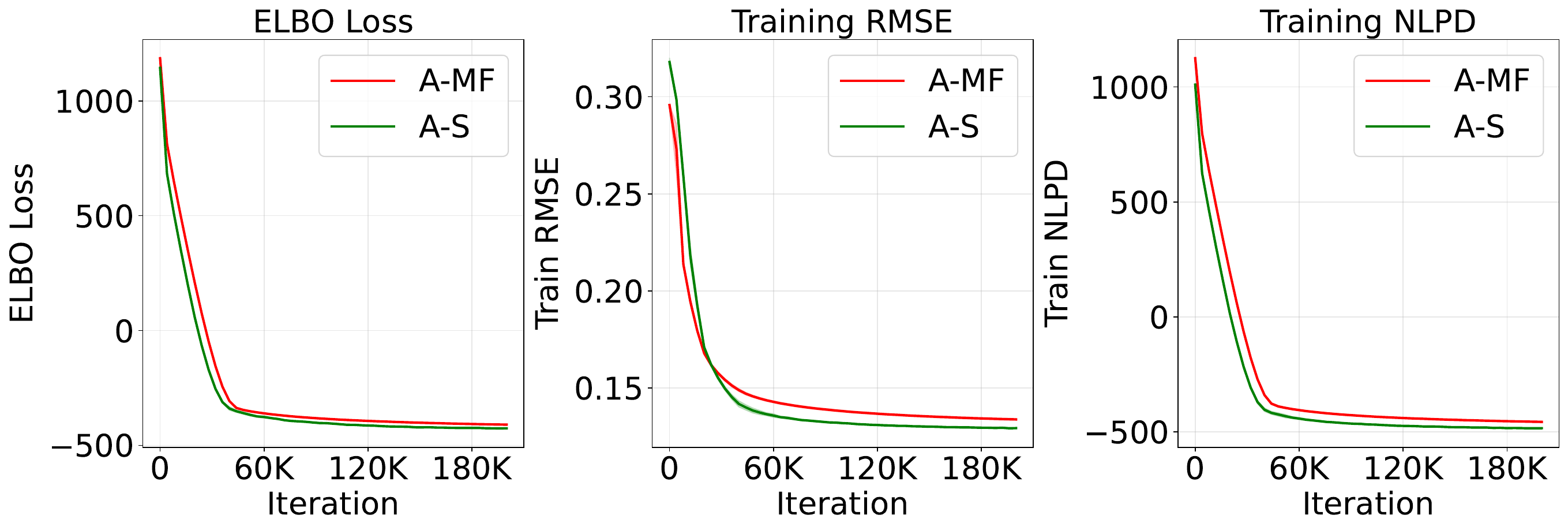}
    \caption{Training dynamics comparison across inference methods on the Fashion-MNIST dataset. 
    The panels show the ELBO, RMSE, and NLPD trajectories over training iterations. 
    } \label{fig:fmnist-metrics}
\end{figure*}